\documentclass{article} 
\usepackage{iclr2027_conference,times}

\usepackage{amsmath,amsfonts,bm}

\def\eqref#1{equation~\ref{#1}}

\def\1{\bm{1}}

\DeclareMathAlphabet{\mathsfit}{\encodingdefault}{\sfdefault}{m}{sl}
\SetMathAlphabet{\mathsfit}{bold}{\encodingdefault}{\sfdefault}{bx}{n}

\usepackage{float}
\usepackage{hyperref}
\usepackage[capitalize,noabbrev]{cleveref}
\usepackage{url}

\usepackage{amsmath,amssymb}
\usepackage{booktabs}
\usepackage{graphicx}
\usepackage{multirow}
\usepackage{tikz}
\usepackage{pgfplots}
\usepackage{xcolor}
\usepackage{colortbl}
\definecolor{dpos}{HTML}{2E7D32}
\definecolor{dneg}{HTML}{C62828}
\definecolor{dzero}{HTML}{8A8A8A}
\definecolor{bestcell}{HTML}{D8E9F5}
\definecolor{secondcell}{HTML}{EDF4FA}
\definecolor{oursrow}{HTML}{FFF7DB}
\definecolor{bandA}{HTML}{E7F0E4}
\definecolor{bandB}{HTML}{F6E0DB}
\definecolor{bandC}{HTML}{EEE8F4}
\definecolor{bandD}{HTML}{ECECEC}
\definecolor{figblue}{HTML}{3D6A8F}
\definecolor{figteal}{HTML}{2E7D6B}
\usepackage{microtype}
\usepackage{pifont}
\usepackage{enumitem}
\usepackage{xcolor}

\usetikzlibrary{arrows.meta,calc,fit,positioning,shapes.geometric}
\usepgfplotslibrary{groupplots}
\pgfplotsset{compat=1.18}

\definecolor{teacherorange}{HTML}{E69F00}
\definecolor{studentblue}{HTML}{0072B2}
\definecolor{workspacepurple}{HTML}{CC79A7}
\definecolor{terminalgreen}{HTML}{009E73}
\definecolor{dangerred}{HTML}{D55E00}
\definecolor{mutedgray}{HTML}{666666}
\definecolor{lightgray}{HTML}{E6E8EB}
\newcommand{\method}{LastOPD}
\newcommand{\codeurl}{https://github.com/Muyiiiii/LastOPD}
\newcommand{\codelink}{\url{\codeurl}}
\newcommand{\rep}{\mathrm{rep}}
\newcommand{\opd}{\mathrm{OPD}}

\newcommand{\T}{\mathrm{T}}
\newcommand{\Smodel}{\mathrm{S}}

\title{LastOPD: Taming Collapse in Latent On-Policy Distillation}

\author{%
Jie Yang$^{1,4,*}$, Zhengyu Fang$^{2,4,*}$, Zelin Xu$^{3,4,*}$, Jiarui Sun$^{4}$, Xiran Fan$^{4}$, \\
\bfseries Junpeng Wang$^{4}$, Liang Wang$^{4}$, Qinghua Liu$^{4,5}$, Yiwei Cai$^{4}$, Yan Zheng$^{4,\dagger}$ \\[4pt]
\mdseries
$^{1}$University of Illinois at Chicago \quad
$^{2}$Case Western Reserve University \quad
$^{3}$University of Florida \\
$^{4}$Visa Research \quad
$^{5}$The Ohio State University \\[2pt]
\texttt{jyang265@uic.edu, yazheng@visa.com}
}

\iclrfinalcopy 
\begin{document}

\maketitle
\lhead{Preprint.}
{\renewcommand{\thefootnote}{\ensuremath{*}}\footnotetext{Equal contribution.\quad$^{\dagger}$Corresponding author.}}

\begin{abstract}
On-policy distillation~(OPD) corrects a student on the responses it writes, but its signal is the teacher's next-token distribution: it tells the student \emph{what} the teacher says but misses \emph{how} it thinks.
Latent supervision promises the missing part by aligning the student's latent states to the teacher's.
Recent methods such as OPRD bring this signal into on-policy distillation.
However, we observe two failures of this recipe when distilling Qwen3-4B and Qwen3-8B into Qwen3-1.7B-Base.
\textbf{Early gain, late collapse}: latent supervision alone lifts MATH-500 accuracy from 25 to 46 in 10 steps, but subsequent training degrades performance down to 11 with no recovery.
\textbf{Better alignment, worse behavior}: although the alignment metric steadily improves throughout this collapse, the most aligned model turns out to be the worst performing.
Further analysis suggests a mismatch in how the latent signal is applied: layers paired by depth play different roles in the two models, so continued alignment may pull the student toward teacher states it cannot understand.
To address this, we propose \method{}, which applies the latent signal only at the last-layer state, the common interface both LM heads read, and only during a 10-step crossfade into token-level OPD.
This keeps the useful part of the latent signal and hands the student to token-level supervision before the collapse sets in.
Extensive experiments show that \method{} improves MATH-500 over token-only OPD by 5.55 and 4.02 points with the 4B and 8B teachers, leads on most held-out datasets, and reaches the final score of token-only OPD in about half the steps.
Code is available at \mbox{\codelink}.

\end{abstract}

\section{Introduction}
\label{sec:intro}

On-policy distillation~(OPD)~\citep{lu2025onpolicydistillation, li2026rethinking, fu2026revisiting} has become a standard post-training recipe, used alongside supervised fine-tuning and reinforcement learning to transfer a teacher's capability into a smaller student \citep{agarwal2024policy,gu2024minillm,yang2025qwen3}.
Unlike off-policy distillation~\citep{hinton2015distilling, kim2016sequence}, which imitates the teacher on text the student did not write itself, OPD lets the student sample its own response and has the teacher score every token it generates, so every correction lands on a context the student actually produces.
Across its variants, which tokens are scored and how they are weighted differ, but the signal they transfer is the same: the teacher's next-token distribution.
However, this token-level distribution supervises only \emph{what} the teacher says and misses \emph{how} it thinks: the computation behind each token stays in the latent states upstream of the LM head and is never scored.

Latent states carry the missing \emph{how}~\citep{hu2026bridging, yang2026observations}: current interpretability studies, including analyses with the Jacobian lens~(J-Lens), show that latent states encode intermediate reasoning steps the output never verbalizes \citep{lindsey2025biology, gurnee2026verbalizable}.
This motivates using the teacher's latent states as an additional source of supervision.
Since FitNets~\citep{romero2015fitnetshintsdeepnets}, latent distillation has built on a single premise: align the student's latent states to the teacher's at chosen layers, and behavioral gains will follow~\citep{sun2019patient,jiao2020tinybert}.
OPRD~\citep{yang2026oprd} carries this premise into the on-policy distillation framework, pairing student and teacher layers by depth and bridging their widths with a projector.

However, it remains unclear whether latent supervision stays beneficial throughout on-policy optimization, and whether better latent alignment reliably translates into better student behavior.
These questions are sharpest when teacher and student differ in depth and width, since nothing guarantees that the paired layers compute the same thing.
To investigate them, we track student performance on MATH-500~\citep{hendrycks2021measuring, lightman2024let} and representational alignment throughout distillation on DAPO-Math-17k~\citep{yu2026dapo}, from Qwen3-4B and Qwen3-8B teachers into a Qwen3-1.7B-Base student.
As illustrated in \cref{fig:intro-draft}, our analysis reveals two key phenomena:
\begin{figure}[!t]
    \centering
    \includegraphics[width=\textwidth]{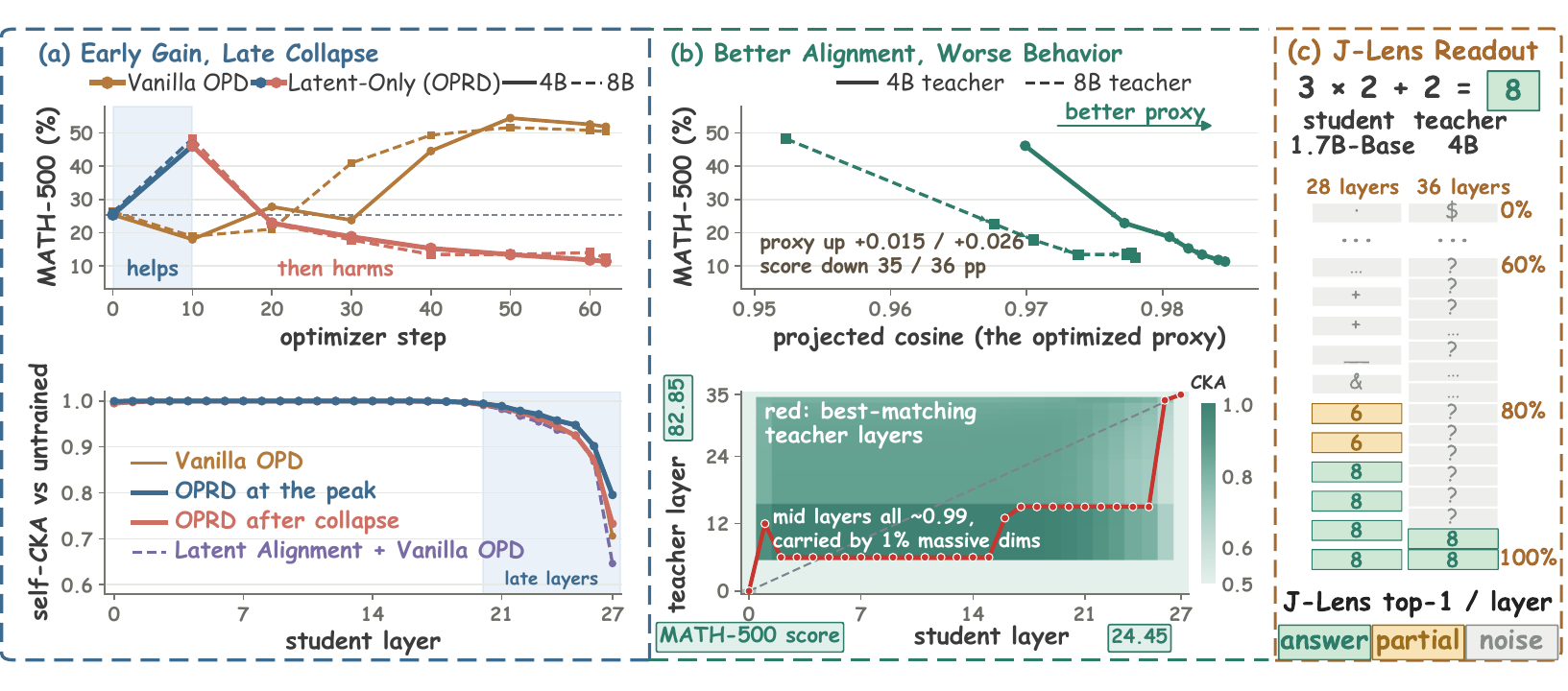}
    \vspace{-15pt}
    \caption{\textbf{Two failure phenomena of latent supervision.}
    \textbf{(a)}~\emph{Early gain, late collapse}: latent-only distillation (OPRD) beats vanilla OPD within 10 steps and then collapses sharply (top), with representational changes concentrated in the student's later layers (bottom).
    \textbf{(b)}~\emph{Better alignment, worse behavior}: during training the optimized projected cosine keeps rising as accuracy falls (top), and before any training the CKA between mid layers is already 0.99, carried by a few massive-activation dimensions (bottom).
    \textbf{(c)}~Per-layer J-lens readout: the student's answer surfaces gradually, whereas the teacher's appears only in its last two layers.}
    \label{fig:intro-draft}
\end{figure}

\begin{itemize}[leftmargin=*]
    \item[\ding{182}]\label{phe:1}
    \textbf{\textcolor{figblue}{Early Gain, Late Collapse}: \emph{latent supervision peaks early and then collapses}~(\cref{fig:intro-draft}a).}
Latent-only distillation~(OPRD) lifts the student from 25.4 to 46.2 on validation MATH-500 within 10 steps, 28 points ahead of vanilla OPD at the same step.
More importantly, continued optimization drives the same run down to 11.3 by the end of training, well below where the student started.
The 8B-teacher experiment shows the same pattern: the early gain later gives way to the collapse.
Moreover, layerwise self centered kernel alignment~(CKA)~\citep{kornblith2019similarity} of the student shows that the main representational changes are concentrated in the student's later layers.
Together, these results suggest that the benefit of latent-only supervision is real but short-lived.

    \item[\ding{183}]\label{phe:2}
    \textbf{\textcolor{figteal}{Better Alignment, Worse Behavior}: \emph{alignment keeps improving while behavior collapses}~(\cref{fig:intro-draft}b).}
Alignment looks high even before it is optimized: the mid layers of the untrained 1.7B student and the 4B teacher already reach a CKA of 0.99, while the two models score 24.5 and 82.9 on MATH-500.
During training, the projected cosine similarity that the all-layer loss optimizes keeps rising after the accuracy peak, from 0.97 to 0.98 with the 4B teacher and from 0.95 to 0.98 with the 8B teacher, while the 4B run's MATH-500 score falls by about 35 points.
This suggests that better alignment under the training objective is not a reliable indicator of better student behavior.
\end{itemize}

Both phenomena motivate a closer examination of how latent supervision is applied in OPD.
OPRD applies latent supervision by pairing student and teacher layers at the same relative depth, on the assumption that the paired layers do the same work.
However, as \cref{fig:intro-draft}c shows, J-Lens readouts~\citep{gurnee2026verbalizable} reveal the teacher's final answer only in its last two layers, whereas the student's answer emerges gradually across layers.
These readout differences suggest a possible source of collapse: continued alignment may force student layers to match teacher states that serve different roles.
This further motivates a central question: \emph{\textbf{How can on-policy distillation keep the gain that latent supervision provides without the collapse?}}

To answer this question, \cref{sec:diagnostics} first examines the collapse more closely, and the analysis leads to \textbf{\method{}}, which tames the collapse in latent on-policy distillation.
Specifically, \method{} aligns the last-layer states before the two LM heads and bridges any width gap with a small trainable MLP.
This places the alignment at a common interface to next-token prediction and needs no intermediate-layer pairing.
On the student's own rollouts, it combines this latent term with reverse top-$k$ token-level loss.
During the first 10 steps, the two terms crossfade: the latent weight decreases linearly as the token weight increases, after which training continues with the token-level loss alone.
In the tested cross-size pairs, \method{} avoids the collapse and improves the final score over both token-only OPD and continued joint supervision.
Our main contributions are summarized as follows:
\begin{itemize}[leftmargin=*]
    \item We identify two phenomena in latent-only OPD: rapid early gains give way to collapse below the student's starting point, while the optimized alignment keeps improving. Diagnostics reveal different teacher--student layerwise readout patterns and similarity scores dominated by a few massive dimensions. Neither masking them nor remapping layers by CKA prevents collapse.
    \item We propose \textbf{\method{}}, which places the latent term on the last-layer states before the LM heads and crossfades into token-level OPD over 10 steps, with no layer pairing or frozen projector.
    \item Extensive experiments on two cross-size Qwen3 pairs show that \method{} improves MATH-500 over token-only OPD by 5.55 and 4.02 points, and improves the mean over eight held-out datasets by 3.93 and 2.10 points, respectively.
\end{itemize}

\section{Related Work}
\label{sec:related}

\paragraph{On-policy distillation.}
On-policy distillation trains the student on its own samples and lets the teacher score every token it generates, which removes the exposure bias of imitating teacher-written text~\citep{agarwal2024policy,gu2024minillm, yang2025qwen3}.
Recent variants keep the recipe and change which tokens are scored and how they are weighted: tail-aware top-$k$ distillation~\citep{huang2026tail} reweights the tokens outside the student's top-$k$ support, and delta distillation~\citep{heo2026policy} replaces the teacher's distribution by its difference from the teacher's base model.
However, whatever the variant, the signal that reaches the student is a single object, the teacher's next-token distribution read out after its LM head, providing the student \emph{what} the teacher says but missing \emph{how} it thinks~\citep{lu2025onpolicydistillation, yang2026oprd}.
\method{} provides the missing part by combining the token-level signal with a latent signal from before the head, kept on only for a short window.

\paragraph{Latent distillation.}
Matching intermediate latent states goes back to the FitNets~\citep{romero2015fitnetshintsdeepnets}, which introduced hint layers as a training target for a thinner student, and later work paired student and teacher layers to align latent states or attention statistics~\citep{sun2019patient,jiao2020tinybert,yang2026glocal,dasgupta2025improving,zhao2025dual}.
Concurrent works bring latent information into on-policy distillation recipes in different ways, including LOPD~\citep{zhang2026latent}, PHF~\citep{li2026phf}, and OPRD~\citep{yang2026oprd}.
However, all of them rest on the same premise, that pulling the student's latent states toward the teacher's brings behavioral gains, without asking whether the target is reachable for a cross-architecture pair or for how long it helps.
We instead analyze why this alignment collapses and where the two models fail to correspond, and the analysis motivates \method{}: a latent term placed at the one state both models share and kept on only for the steps in which it helps.

\section{Preliminaries}
\label{sec:prelim}

\paragraph{On-Policy Distillation.}
Let $\pi_\theta$ be the student and $\pi_\T$ a frozen teacher that share a tokenizer and vocabulary $\mathcal V$.
For each prompt $x$ the student samples its own response $\hat y\sim\pi_\theta(\cdot\mid x)$, and both models are evaluated on the same prefixes $s_t=(x,\hat y_{<t})$, so every correction lands on a context the student actually produces \citep{agarwal2024policy,gu2024minillm}.
We use the reverse top-$k$ form of \citet{yang2025qwen3}.
Let $p_{\theta,t}$ and $p_{\T,t}$ be the two models' next-token distributions at $s_t$.
Let $\mathcal V_{k,t}$ be the $k$ tokens the student ranks highest, and $R_{\mathcal V_{k,t}}(p)$ the restriction of a distribution to this support, renormalized.
The token-level objective is as follows:
\begin{equation}
    \mathcal L_{\opd}
    = \frac{1}{|\mathcal M|}\sum_{t\in\mathcal M}
      D_{\mathrm{KL}}\!\left(
        R_{\mathcal V_{k,t}}(p_{\theta,t})
        \,\middle\|\,
        R_{\mathcal V_{k,t}}(p_{\T,t})
      \right),
    \label{eq:opd}
\end{equation}
where $\mathcal M$ is the set of response positions and $k=16$ throughout.
Training with \cref{eq:opd} alone is what we call token-only OPD in our tables.
Whatever the choice of support and weighting, the signal that reaches the student is the teacher's next-token distribution, read out after the teacher's LM head.

\paragraph{Latent supervision.}
Latent distillation adds a second supervision signal from before the LM head.
Since FitNets~\citep{romero2015fitnetshintsdeepnets}, it has rested on a single premise: pick a student layer and a teacher layer, pull the student's state toward the teacher's, and behavioral gains will follow~\citep{sun2019patient,jiao2020tinybert,wang2020minilm}.
Let $z^\Smodel_t\in\mathbb R^{d_\Smodel}$ and $z^\T_t\in\mathbb R^{d_\T}$ be the two states at position $t$, and let $g_\psi$ be a projector that bridges the two widths.
With $\nu(z)=z/(\lVert z\rVert_2+\epsilon)$ and stop-gradient $\operatorname{sg}$ on the teacher side, the generic latent loss is
\begin{equation}
    \mathcal L_{\rep}
    = \frac{1}{|\mathcal M|\,d_\T}\sum_{t\in\mathcal M}
      \left\lVert
      \nu\!\left(g_\psi(z^\Smodel_t)\right)
      -
      \operatorname{sg}\!\left[\nu(z^\T_t)\right]
      \right\rVert_2^2 .
    \label{eq:latent}
\end{equation}
OPRD~\citep{yang2026oprd} brings this premise into on-policy distillation: \cref{eq:latent} is computed on the student's own rollouts at every layer, with layers paired by relative depth and widths bridged by a frozen low-rank projector.
They name this cross-architecture instantiation OPRD-Bridge, and its same-architecture form, which needs no projector, OPRD-Vanilla.

\section{Why Latent Supervision Collapses}
\label{sec:diagnostics}

\cref{sec:intro} showed what happens when latent supervision stays on, and this section asks why.
Unless stated otherwise, all measurements use the Qwen3-4B teacher and the Qwen3-1.7B-Base student, and MATH-500 is the test mean over eight samples, while training curves follow a lighter validation protocol within about two points of the test score.

\paragraph{What the collapse looks like.}

\begin{figure}[t]
\centering
\includegraphics[width=\textwidth]{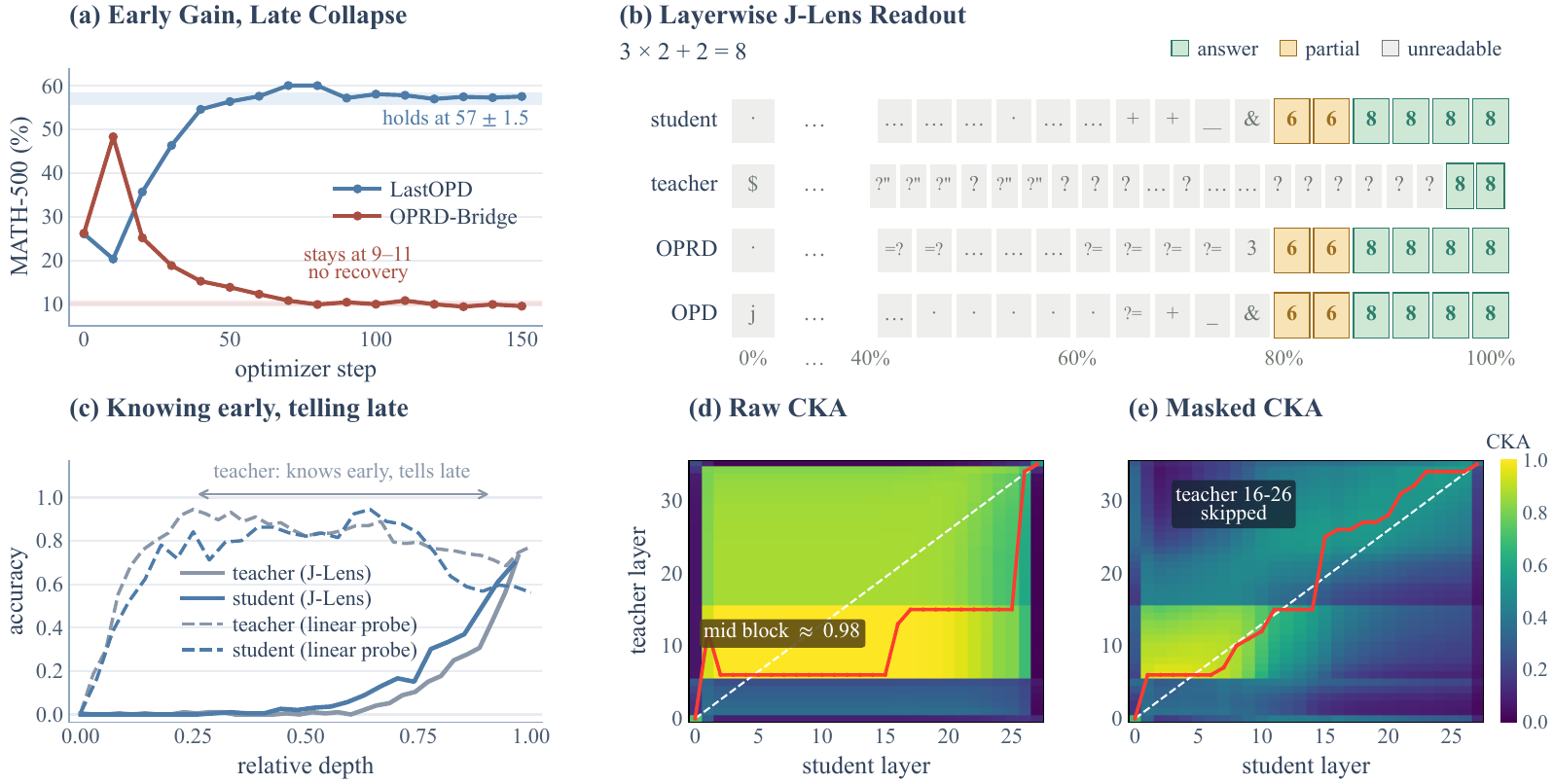}
\vspace{-6pt}
\caption{\textbf{What the collapse looks like, and where the models truly correspond.}
\textbf{(a)}~Validation MATH-500 for OPRD-Bridge and \method{} runs extended to 150 steps.
\textbf{(b)}~J-Lens top-1 readouts on ``$3*2+2=$''. The uninformative 0--40\% interval is compressed.
\textbf{(c)}~Per-layer readout agreement (solid lines) and running-value probe accuracy (dashed lines).
\textbf{(d)}~Raw CKA over all layer pairs.
\textbf{(e)}~The same map with massive dimensions removed.
Details in Appendix~\ref{app:diag}.}
\label{fig:diag-anatomy}
\end{figure}

To examine whether the collapse persists with longer training, we extend the OPRD-Bridge run from 62 to 150 steps.
As \cref{fig:diag-anatomy}a shows, the validation score stays between 9.4 and 10.8 from step 70 onward, without any recovery.
The collapsed student is not silent either.
On MATH-500, roughly two thirds of its answers are short, boxed, and wrong, and fewer than one in ten loop (Appendix~\ref{app:collapsed-says}).
So the degradation under continued latent-only supervision is persistent and consists mostly of wrong answers.
To understand this failure, we next examine whether the layers OPRD pairs by relative depth play comparable roles in the teacher and student.

\begin{figure}[t]
\centering
\includegraphics[width=\textwidth]{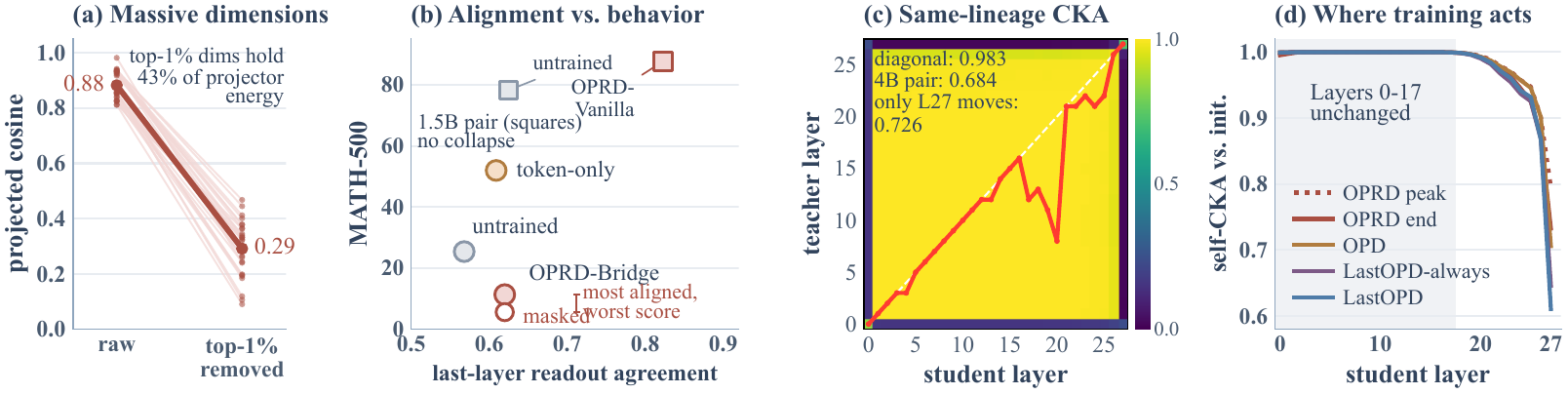}
\vspace{-15pt}
\caption{\textbf{Inflated alignment, and the difference that matters.}
\textbf{(a)}~Projected cosine per layer pair, with and without the massive dimensions.
\textbf{(b)}~Validation MATH-500 at the final step against last-layer readout agreement. Hollow: masked retrain. Squares: same-lineage pair.
\textbf{(c)}~Same-lineage CKA with the ridge (red).
\textbf{(d)}~Self-CKA against initialization.
}
\label{fig:diag-correspond}
\end{figure}

\paragraph{When each model knows and tells.}

To compare the paired layers, we use J-Lens~\citep{gurnee2026verbalizable} to read out next-token predictions at each layer.
\cref{fig:diag-anatomy}b shows the readouts on ``$3*2+2=$'': the student reads out the intermediate product 6 at two layers and then the answer 8 from the next layer on, while the teacher shows nothing readable until the answer appears at its last two layers.
\cref{fig:diag-anatomy}c~(solid lines) extends this comparison to 192 mid-generation positions, scoring agreement with the token each model eventually emits: the teacher's agreement stays near zero until 63\% of its depth and then jumps, while the untrained student's rises gradually from layer 12.
But a readout only shows what a layer is willing to say, not everything it knows.
To probe the information already present, we further train a linear probe on a separate arithmetic task to predict partial sums from each layer's state (protocol in Appendix~\ref{app:proto-probe}).
In \cref{fig:diag-anatomy}c (dashed lines), the teacher's probe reaches 94\% accuracy by layer 9, long before its readout moves.
The student's probe passes 80\% by layer 7 and peaks at layer 18, where its readout is already climbing.
Together, these depth profiles suggest a teacher that knows early and tells late, and a student that tells as it goes.
These differences suggest that layers at the same relative depth may play different readout roles.
If depth alone does not tell us which layers correspond, similarity might, and we check that next.

\paragraph{Can similarity guide layer pairing?}

To test whether similarity can guide layer pairing, we compute CKA~\citep{kornblith2019similarity} for every student--teacher layer pair and plot the result in \cref{fig:diag-anatomy}d, where the red line selects the most similar teacher layer for each student layer and the white diagonal marks pairing by relative depth.
For the 4B pair, the middle block averages 0.98 with little contrast among candidate pairs, and the red line concentrates on a few teacher layers.
By contrast, the same-lineage JustRL-1.5B and R1-Distill-1.5B pair in \cref{fig:diag-correspond}c shows a clear diagonal at 0.983 against 0.684 for the 4B pair, and its latent-only run in \cref{fig:diag-correspond}b does not collapse.
To understand what drives the high scores in the 4B map, we inspect the underlying activations and find massive activations~\citep{sun2024massive} in all five models surveyed: a few dimensions whose magnitude is far above the median~(Appendix~\ref{app:proto-massive}).
These dimensions dominate raw CKA and carry 43\% of the frozen projector's energy: as \cref{fig:diag-correspond}a shows, masking them lowers the step-one projected cosine from 0.88 to 0.29.
In \cref{fig:diag-anatomy}e, the same mask moves the CKA ridge closer to the diagonal, though teacher layers 16--26 are never selected along the ridge and the last-layer match remains weak at 0.38.
Together, these results show that measured similarity is strongly influenced by a few massive dimensions, which limits its use as evidence of functional correspondence.
So the question is whether masking these dimensions or changing the layer pairing can prevent the collapse.

\paragraph{Do masking and remapping prevent the collapse?}

If the inflated similarity is what misleads the latent loss, removing the massive dimensions from the loss should improve task performance.
We test this by retraining OPRD-Bridge with the massive dimensions masked out of its latent loss, and \cref{fig:diag-correspond}b shows the opposite: its final validation MATH-500 score falls from 11.33 to 5.60.
We also test whether replacing depth-based pairing with the raw CKA ridge helps, but this run repeats the same arc and peaks near 50 at step 10 before collapsing by step 20~(Appendix~\ref{app:ridge}).
Neither change prevents the collapse, and what the two share is that both still supervise only latent states and never the student's next-token predictions.
The one change that does hold the student adds exactly that: in \cref{tab:transfer}, adding token-level OPD to OPRD-Bridge raises its official MATH-500 score from 12.12 to 53.70 with the latent term still active, close to the 53.40 of token-only OPD alone.
So kept on throughout, the latent term adds nothing on top of the token term.
Based on these observations, we get a hypothesis that the teacher's state mixes a part the student can understand with a part it cannot: the first is matched within the first steps and gives the early gain, while continued matching pulls the student toward the second, which impairs predictions and adds nothing beyond token-level OPD.

\section{Methodology}
\label{sec:method}

\begin{figure}[t]
    \centering
    \includegraphics[width=\textwidth]{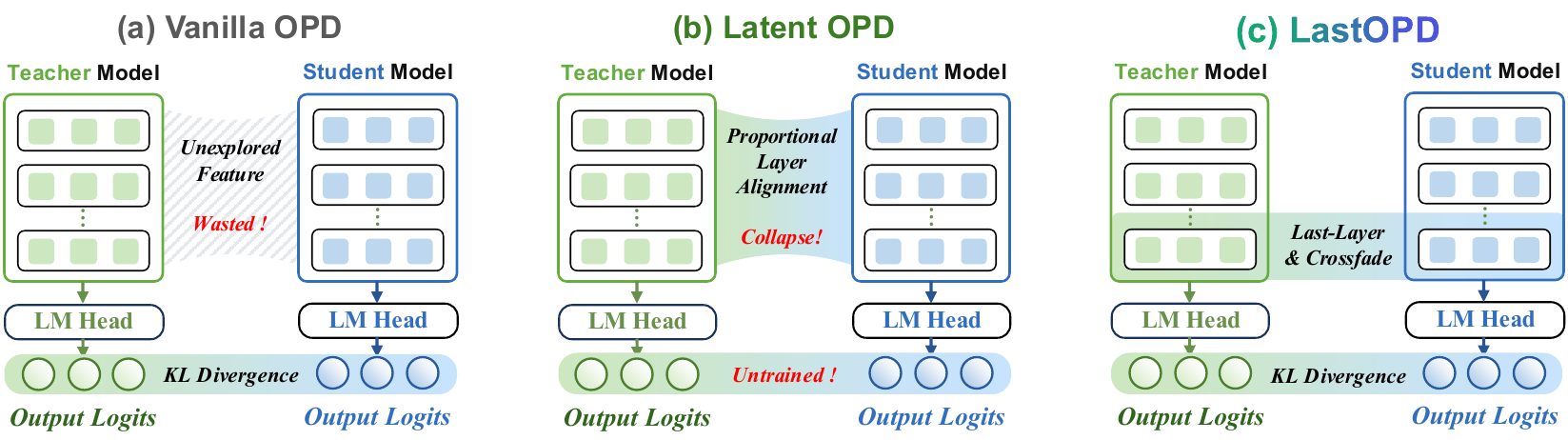}
    \vspace{-15pt}
    \caption{\textbf{From output-only and layerwise latent OPD to \method{}.}
    \textbf{(a)} Vanilla OPD matches next-token distributions and leaves the teacher's latent states unused.
    \textbf{(b)} Layerwise latent OPD aligns depth-paired layers for the whole run and provides no direct token-level supervision.
    \textbf{(c)} \method{} aligns only the last-layer state and crossfades to token-level OPD within the first 10 steps.}
    \label{fig:overview}
\end{figure}

\method{} keeps the useful part of the latent signal without the collapse that follows, using only the student's own rollouts.
As \cref{fig:overview} illustrates, it differs from output-only OPD~(a) and layerwise latent OPD~(b) in where and when the latent signal is applied.

\paragraph{Where: the last-layer state.}
Both LM heads read the post-normalization last-layer states $z^\Smodel_t$ and $z^\T_t$ to predict the next token, so these states provide a common interface to next-token prediction across models of different depths and widths.
\method{} applies \cref{eq:latent} at this state only, with a small trainable MLP $g_\psi:\mathbb R^{d_\Smodel}\to\mathbb R^{d_\T}$ that bridges any width gap and reduces to the identity when the widths match.
Gradients update the student $\theta$ and the adapter $\psi$ while the teacher stays frozen, and no layer pairing or frozen projector is needed.
The token term is the reverse top-$k$ OPD of \cref{eq:opd} on the same rollouts, which scores the student's own top-$k$ candidates.

\paragraph{When: a 10-step crossfade.}
\label{sec:method-crossfade}
\cref{sec:diagnostics} suggests that the useful part of the latent signal arrives within the first steps and that continued latent-only training leads to collapse.
If so, the latent weight should fall with training rather than stay constant, and an early handoff to token-level OPD should beat keeping the term on throughout.
So over a window of $T_w$ steps the latent weight falls linearly to zero while the token weight rises linearly to one:
\begin{equation}
    \mathcal L_t=\alpha(t)\,\mathcal L_{\opd}+\lambda_{\rep}\,\beta(t)\,\mathcal L_{\rep},
    \qquad
    \alpha(t)=\min\!\left(1,\,\tfrac{t}{T_w}\right),
    \qquad
    \beta(t)=\max\!\left(0,\,1-\tfrac{t}{T_w}\right).
    \label{eq:crossfade}
\end{equation}
At step $0$ the objective is the latent term alone, and from step $T_w$ on it is exactly token-only OPD.
We set $T_w=10$, the step at which latent-only distillation peaks in \cref{sec:diagnostics}.
The projector is discarded at inference, and the base weight $\lambda_{\rep}$ and the remaining overhead are described in Appendix~\ref{app:setup}.

\section{Experiments}
\label{sec:experiments}

\subsection{Experimental Setup}
\label{sec:setup}

\paragraph{Models.}
Teachers are Qwen3-4B, Qwen3-8B~\citep{yang2025qwen3}, and JustRL-1.5B~\citep{he2025justrl}, and students are Qwen3-1.7B-Base~\citep{yang2025qwen3} and R1-Distill-1.5B~\citep{guo2025deepseek}.
The Qwen3 teachers distill into Qwen3-1.7B-Base across depth and width, and JustRL-1.5B into R1-Distill-1.5B, a same-lineage control of one architecture.

\paragraph{Protocol.}
Following the settings of OPRD~\citep{yang2026oprd}, every method trains for 62 optimizer steps on DAPO-Math-17k~\citep{yu2026dapo}, about 7.9k rollouts, in Qwen3's non-thinking mode, with the same prompts, batches, and optimizer for every method.
Every cell is a single run taken at its final checkpoint, and full hyperparameters are in Appendix~\ref{app:setup}.

\paragraph{Evaluation.}
We evaluate on eight math datasets held out from training: MATH-500~\citep{hendrycks2021measuring, lightman2024let}, AIME24, AIME25, AMC23,  Minerva~\citep{lewkowycz2022solving}, OlympiadBench~\citep{he2024olympiadbench}, AIMO~\citep{ai-mathematical-olympiad-prize}, and GSM8K~\citep{cobbe2021training}.

\paragraph{Baselines.}
We compare against token-only OPD~\citep{lu2025onpolicydistillation} and OPRD-Bridge~\citep{yang2026oprd} with and without the token-level loss term, plus OPRD-Vanilla on the same-lineage pair.
\method{}-always keeps both terms of \method{} at constant weight for the whole run~(Appendix~\ref{app:setup}).

\subsection{Overall Comparison}
\label{sec:res-main}

\begin{table}[t]
\caption{\textbf{Main comparison across three teacher--student pairs.} Qwen3-4B and Qwen3-8B distill into Qwen3-1.7B-Base, and JustRL-1.5B into R1-Distill-1.5B, whose latent-only recipe is OPRD-Vanilla. Both Mean columns average all eight datasets, with the acc@1 of GSM8K entering each. \colorbox{bestcell}{\textbf{Bold}} marks the best trained method per column and block and \colorbox{secondcell}{\underline{underline}} the second.}
\label{tab:transfer}
\centering
\scriptsize
\setlength{\tabcolsep}{3pt}
\renewcommand{\arraystretch}{1.15}
\resizebox{\textwidth}{!}{%
\begin{tabular}{l|*{15}{c}|cc}
\toprule
\multirow{2}{*}{\textbf{Method}} & \multicolumn{2}{c}{MATH-500} & \multicolumn{2}{c}{AIME24} & \multicolumn{2}{c}{AIME25} & \multicolumn{2}{c}{AMC23} & \multicolumn{2}{c}{Minerva} & \multicolumn{2}{c}{OlympiadBench} & \multicolumn{2}{c}{AIMO} & GSM8K & \multicolumn{2}{|c}{Mean} \\
 & avg@8 & best@8 & avg@16 & best@16 & avg@16 & best@16 & avg@8 & best@8 & avg@4 & best@4 & avg@4 & best@4 & avg@16 & best@16 & acc@1 & avg & best \\
\midrule
\rowcolor{bandA}\multicolumn{18}{c}{\textbf{Qwen3-4B $\rightarrow$ Qwen3-1.7B-Base}} \\
Teacher~(Qwen3-4B) & 82.85 & 94.2 & 22.08 & 56.7 & 23.54 & 46.7 & 69.06 & 90.0 & 31.34 & 38.2 & 47.48 & 60.3 & 60.47 & 84.3 & 91.51 & 53.54 & 70.2 \\
Student~(Qwen3-1.7B-Base) & 24.45 & 72.2 & 2.71 & 10.0 & 0.62 & 10.0 & 18.44 & 50.0 & 6.80 & 18.4 & 9.33 & 16.6 & 9.19 & 44.6 & 38.51 & 13.76 & 32.5 \\
\midrule
Token-only OPD & 53.40 & 81.2 & \cellcolor{bestcell}\textbf{9.17} & \cellcolor{secondcell}\underline{23.3} & \cellcolor{bestcell}\textbf{5.21} & \cellcolor{bestcell}\textbf{20.0} & 27.50 & 60.0 & 13.14 & 24.6 & 20.78 & 36.9 & 26.58 & \cellcolor{secondcell}\underline{63.9} & 67.78 & 27.95 & 47.2 \\
OPRD-Bridge~(latent-only) & 12.12 & 33.8 & 0.21 & 3.3 & 0.00 & 0.0 & 3.12 & 20.0 & 7.35 & 16.9 & 2.63 & 7.4 & 3.69 & 28.9 & 8.72 & 4.73 & 14.9 \\
OPRD-Bridge + token & 53.70 & 83.0 & \cellcolor{secondcell}\underline{8.33} & \cellcolor{bestcell}\textbf{26.7} & \cellcolor{secondcell}\underline{3.96} & \cellcolor{bestcell}\textbf{20.0} & \cellcolor{secondcell}\underline{33.44} & 62.5 & 13.33 & 26.8 & \cellcolor{secondcell}\underline{22.89} & \cellcolor{bestcell}\textbf{40.9} & 26.43 & 62.6 & \cellcolor{secondcell}\underline{69.67} & \cellcolor{secondcell}\underline{28.97} & 49.0 \\
\cellcolor{oursrow}\textbf{\method{}-always~(no fade)} & \cellcolor{secondcell}\underline{53.77} & \cellcolor{bestcell}\textbf{84.2} & 8.12 & \cellcolor{secondcell}\underline{23.3} & 3.75 & \cellcolor{secondcell}\underline{16.7} & 30.63 & \cellcolor{bestcell}\textbf{72.5} & \cellcolor{secondcell}\underline{15.07} & \cellcolor{secondcell}\underline{29.0} & 22.30 & \cellcolor{secondcell}\underline{39.9} & \cellcolor{secondcell}\underline{28.54} & \cellcolor{secondcell}\underline{63.9} & 68.84 & 28.88 & \cellcolor{secondcell}\underline{49.8} \\
\cellcolor{oursrow}\textbf{\method{}~(ours)} & \cellcolor{bestcell}\textbf{58.95} & \cellcolor{secondcell}\underline{83.6} & 7.50 & \cellcolor{secondcell}\underline{23.3} & \cellcolor{secondcell}\underline{3.96} & \cellcolor{bestcell}\textbf{20.0} & \cellcolor{bestcell}\textbf{35.00} & \cellcolor{secondcell}\underline{70.0} & \cellcolor{bestcell}\textbf{17.74} & \cellcolor{bestcell}\textbf{31.6} & \cellcolor{bestcell}\textbf{24.26} & 38.8 & \cellcolor{bestcell}\textbf{31.78} & \cellcolor{bestcell}\textbf{66.3} & \cellcolor{bestcell}\textbf{75.82} & \cellcolor{bestcell}\textbf{31.88} & \cellcolor{bestcell}\textbf{51.2} \\
\midrule
\rowcolor{bandB}\multicolumn{18}{c}{\textbf{Qwen3-8B $\rightarrow$ Qwen3-1.7B-Base}} \\
Teacher~(Qwen3-8B) & 83.38 & 94.4 & 21.25 & 46.7 & 20.62 & 46.7 & 68.12 & 92.5 & 28.31 & 36.0 & 47.85 & 60.4 & 62.42 & 86.8 & 93.10 & 53.13 & 69.6 \\
Student~(Qwen3-1.7B-Base) & 24.45 & 72.2 & 2.71 & 10.0 & 0.62 & 10.0 & 18.44 & 50.0 & 6.80 & 18.4 & 9.33 & 16.6 & 9.19 & 44.6 & 38.51 & 13.76 & 32.5 \\
\midrule
Token-only OPD & \cellcolor{secondcell}\underline{49.43} & 81.8 & \cellcolor{secondcell}\underline{7.50} & 16.7 & \cellcolor{bestcell}\textbf{4.79} & \cellcolor{secondcell}\underline{20.0} & \cellcolor{secondcell}\underline{27.50} & \cellcolor{secondcell}\underline{62.5} & \cellcolor{secondcell}\underline{12.78} & 25.4 & 20.00 & \cellcolor{secondcell}\underline{37.0} & 23.12 & 59.0 & \cellcolor{secondcell}\underline{65.13} & \cellcolor{secondcell}\underline{26.28} & 45.9 \\
OPRD-Bridge~(latent-only) & 12.32 & 32.2 & 0.62 & 10.0 & 0.21 & 3.3 & 3.44 & 15.0 & 5.97 & 16.2 & 2.37 & 6.8 & 4.82 & 25.3 & 11.22 & 5.12 & 15.0 \\
OPRD-Bridge + token & 47.05 & 80.8 & 5.42 & 16.7 & 3.96 & \cellcolor{secondcell}\underline{20.0} & 26.56 & 55.0 & 11.76 & \cellcolor{secondcell}\underline{26.1} & 19.89 & 35.9 & 23.42 & \cellcolor{secondcell}\underline{62.6} & 63.46 & 25.19 & 45.1 \\
\cellcolor{oursrow}\textbf{\method{}-always~(no fade)} & 48.95 & \cellcolor{bestcell}\textbf{83.6} & 5.00 & \cellcolor{secondcell}\underline{20.0} & \cellcolor{secondcell}\underline{4.58} & \cellcolor{bestcell}\textbf{23.3} & 26.25 & \cellcolor{bestcell}\textbf{65.0} & 9.47 & 24.6 & \cellcolor{secondcell}\underline{20.19} & 36.7 & \cellcolor{secondcell}\underline{24.32} & 53.0 & 63.00 & 25.22 & \cellcolor{secondcell}\underline{46.2} \\
\cellcolor{oursrow}\textbf{\method{}~(ours)} & \cellcolor{bestcell}\textbf{53.45} & \cellcolor{secondcell}\underline{82.4} & \cellcolor{bestcell}\textbf{7.92} & \cellcolor{bestcell}\textbf{30.0} & \cellcolor{secondcell}\underline{4.58} & \cellcolor{secondcell}\underline{20.0} & \cellcolor{bestcell}\textbf{27.81} & \cellcolor{bestcell}\textbf{65.0} & \cellcolor{bestcell}\textbf{13.05} & \cellcolor{bestcell}\textbf{28.3} & \cellcolor{bestcell}\textbf{21.19} & \cellcolor{bestcell}\textbf{38.5} & \cellcolor{bestcell}\textbf{27.41} & \cellcolor{bestcell}\textbf{68.7} & \cellcolor{bestcell}\textbf{71.65} & \cellcolor{bestcell}\textbf{28.38} & \cellcolor{bestcell}\textbf{50.6} \\
\midrule
\rowcolor{bandC}\multicolumn{18}{c}{\textbf{JustRL-1.5B $\rightarrow$ R1-Distill-1.5B}} \\
Teacher~(JustRL-1.5B) & 89.55 & 94.8 & 53.96 & 80.0 & 37.92 & 56.7 & 90.62 & 97.5 & 33.64 & 41.9 & 56.44 & 64.4 & 82.08 & 95.2 & 84.91 & 66.14 & 76.9 \\
Student~(R1-Distill-1.5B) & 83.67 & 94.0 & 29.58 & 66.7 & 23.12 & 43.3 & 71.56 & 92.5 & 29.04 & 40.4 & 44.11 & 57.0 & 63.78 & 89.2 & 80.14 & 53.12 & 70.4 \\
\midrule
Token-only OPD & 86.02 & \cellcolor{bestcell}\textbf{94.6} & \cellcolor{bestcell}\textbf{49.79} & \cellcolor{bestcell}\textbf{80.0} & \cellcolor{bestcell}\textbf{35.42} & \cellcolor{bestcell}\textbf{56.7} & 85.00 & \cellcolor{bestcell}\textbf{97.5} & \cellcolor{bestcell}\textbf{32.63} & 40.1 & 51.85 & 61.9 & 72.89 & 89.2 & 85.44 & 62.38 & \cellcolor{secondcell}\underline{75.7} \\
OPRD-Vanilla~(latent-only) & \cellcolor{bestcell}\textbf{87.22} & \cellcolor{secondcell}\underline{94.2} & 47.71 & \cellcolor{secondcell}\underline{76.7} & 33.96 & \cellcolor{secondcell}\underline{53.3} & \cellcolor{secondcell}\underline{86.56} & \cellcolor{bestcell}\textbf{97.5} & 31.07 & 40.1 & \cellcolor{bestcell}\textbf{56.64} & \cellcolor{bestcell}\textbf{65.6} & \cellcolor{secondcell}\underline{78.54} & \cellcolor{secondcell}\underline{92.8} & 86.20 & \cellcolor{bestcell}\textbf{63.49} & \cellcolor{bestcell}\textbf{75.8} \\
OPRD-Vanilla + token & \cellcolor{secondcell}\underline{86.35} & 93.8 & 41.25 & 73.3 & 33.33 & 50.0 & 81.56 & 92.5 & \cellcolor{secondcell}\underline{32.17} & 40.1 & \cellcolor{secondcell}\underline{52.89} & 60.4 & 75.15 & 91.6 & \cellcolor{secondcell}\underline{88.70} & 61.43 & 73.8 \\
\cellcolor{oursrow}\textbf{\method{}-always~(no fade)} & 84.47 & 93.8 & \cellcolor{secondcell}\underline{49.58} & \cellcolor{bestcell}\textbf{80.0} & 33.33 & 46.7 & 85.94 & \cellcolor{secondcell}\underline{95.0} & 31.53 & \cellcolor{secondcell}\underline{40.4} & 52.30 & 62.3 & \cellcolor{bestcell}\textbf{78.69} & \cellcolor{bestcell}\textbf{95.2} & \cellcolor{bestcell}\textbf{89.08} & \cellcolor{secondcell}\underline{63.12} & 75.3 \\
\cellcolor{oursrow}\textbf{\method{}~(ours)} & 83.42 & \cellcolor{bestcell}\textbf{94.6} & 48.12 & 73.3 & \cellcolor{secondcell}\underline{34.38} & 50.0 & \cellcolor{bestcell}\textbf{86.88} & \cellcolor{secondcell}\underline{95.0} & 31.80 & \cellcolor{bestcell}\textbf{42.3} & 52.19 & \cellcolor{secondcell}\underline{62.5} & 77.79 & \cellcolor{secondcell}\underline{92.8} & \cellcolor{secondcell}\underline{88.70} & 62.91 & 74.9 \\
\bottomrule
\end{tabular}}
\end{table}

\cref{tab:transfer} compares \method{} with the baselines on the three teacher--student pairs.
Each block holds the frozen teacher, the untrained student, token-only OPD, the two OPRD recipes, and the two schedules of our latent term, so that every gain can be traced to one change.
We summarize our observations~(\textbf{Obs.}) as follows:

\textbf{Obs.~\ding{182}: The gains extend across teacher sizes and held-out datasets.}
With the Qwen3-4B and Qwen3-8B teachers, \method{} reaches 58.95 and 53.45 on MATH-500, exceeding token-only OPD by 5.55 and 4.02 points, respectively.
Keeping the same last-layer term on throughout, \method{}-always reaches 53.77 and 48.95, so the early crossfade improves performance by 5.18 and 4.50 points over continued joint supervision.
OPRD-Bridge + token, which adds every-layer latent supervision to the token term, ends at 53.70 and 47.05, close to token-only OPD with the 4B teacher and below it with the 8B teacher.
Moreover, the gain is not confined to MATH-500: \method{} leads token-only OPD on five and six of the seven other datasets, by up to 8.0 and 6.5 points, and by 3.93 and 2.10 points on the eight-dataset mean.
The best@$N$ columns tell the same story: \method{} has the highest best@$N$ mean in both blocks, 51.2 and 50.6 against 47.2 and 45.9 for token-only OPD.
OPRD-Bridge alone collapses to 12.12 and 12.32, below the untrained student, as \cref{sec:diagnostics} predicts.
The same latent signal thus collapses the student when applied at every layer throughout and lifts it when confined to the last layer and switched off in time, which answers the question of \cref{sec:intro}.

\textbf{Obs.~\ding{183}: Continued latent supervision performs better on the same-lineage pair.}
On MATH-500 for the JustRL-1.5B and R1-Distill-1.5B pair, latent-only OPRD-Vanilla achieves the highest score of 87.22 above the student's 83.67 and token-only OPD's 86.02.
Keeping the last-layer latent term active also yields a higher score than crossfading, with \method{}-always reaching 84.47 against 83.42 for \method{}.
\method{} itself stays within 0.58 points of the best method on the eight-dataset mean, so a fixed 10-step fade shared across all pairs comes at a modest cost.
This is consistent with our hypothesis and with the layer map of \cref{fig:diag-correspond}c, where this pair's diagonal CKA reaches 0.983 against 0.684 for the 4B pair: when the layers already correspond, the teacher's state may largely consist of what the student can understand and the latent term remains useful throughout training.

\begin{figure}[t]
\centering
\includegraphics[width=\textwidth]{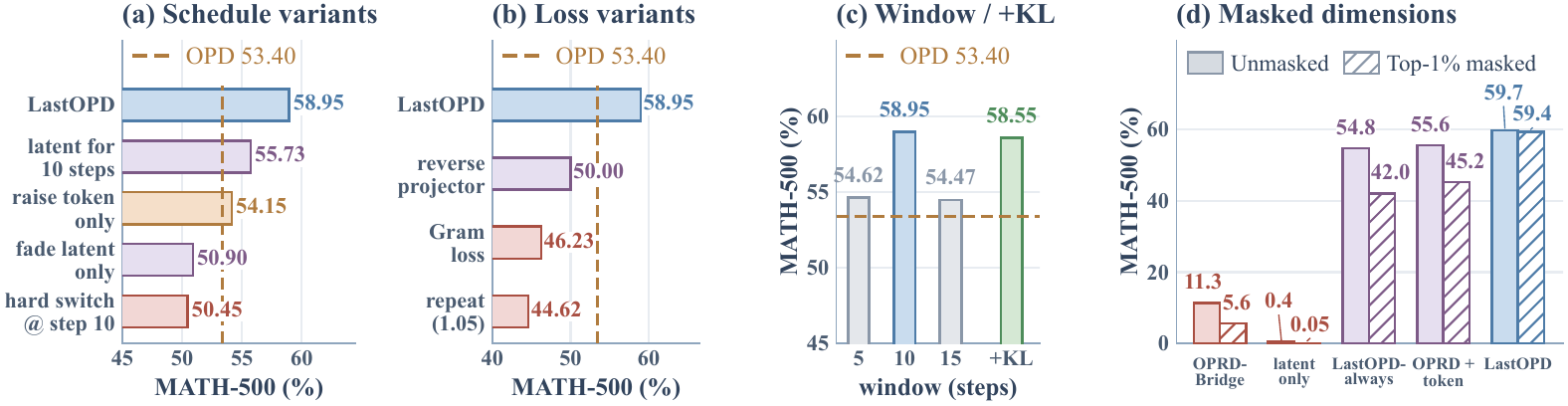}
\vspace{-15pt}
\caption{\textbf{Ablation and sensitivity study} (Qwen3-4B teacher, official MATH-500 except (d)). \textbf{(a)}~Schedule variants, with token-only OPD dashed. \textbf{(b)}~Alternative forms of the latent loss. \textbf{(c)}~Crossfade window and the reference KL term. \textbf{(d)}~The always-on recipes and \method{} with and without the massive-activation mask, on validation MATH-500 at the final step.}
\vspace{-5pt}
\label{fig:res-ablation}
\end{figure}

\subsection{Ablation and Sensitivity Study}
\label{sec:res-ablation}

\cref{fig:res-ablation} takes \method{} apart with the Qwen3-4B teacher, and \cref{tab:ablation} in Appendix~\ref{app:ablation-table} lists every run.
The ablation removes one design choice at a time: \textbf{(a)}~replaces the crossfade with a hard switch or with only one of its two halves, and \textbf{(b)}~swaps the latent loss for a reversed projector or a Gram-matrix loss.
The sensitivity study then \textbf{(c)}~varies the window length $T_w$ and adds a reference KL term.
As a check on the diagnosis, \textbf{(d)}~applies the massive-activation mask of \cref{sec:diagnostics} to every recipe that keeps the latent term on and to \method{}.
We observe:

\textbf{Obs.~\ding{184}: The two halves of the crossfade work best together.}
As \cref{fig:res-ablation}a shows, a hard switch at step 10 drops MATH-500 from 58.95 to 50.45, below token-only OPD, so the same two losses applied one after the other are worse than never using the latent term at all.
Keeping only one half of the crossfade does not recover the full gain: raising the token weight from 0 to 1 without any latent term reaches 54.15, and fading the latent weight over a constant token weight reaches 50.90, below token-only OPD.
What does recover it is running the two terms at the same time: holding the latent weight constant for 10 steps while the token weight rises reaches 55.73, and letting the latent weight fade over those same steps adds the final 3.22 points.
So the rising token weight alone cannot explain the full gain, which depends on introducing latent supervision early and reducing its weight gradually.

\textbf{Obs.~\ding{185}: None of the five variants improves on the full recipe.}
\cref{fig:res-ablation}b and c change the recipe in three directions around the full method: the window, the form of the latent loss, and an added regularizer.
The smallest change already costs over 4 points: as \cref{fig:res-ablation}c shows, windows of 5 and 15 steps reach 54.62 and 54.47, close to token-only OPD, whereas the 10 steps of the full method sit at the step where latent-only distillation peaks in \cref{sec:diagnostics}.
A larger change costs more, as \cref{fig:res-ablation}b shows.
A reversed projector that maps the teacher's state into the student's space gives 50.00, and a Gram-matrix loss that matches the two models' $T\times T$ cosine-similarity matrices over response positions without any projector gives 46.23, both below token-only OPD.
Adding a regularizer does not help either: a KL term to the initial policy, which constrains how far the student moves from its starting point and rises with the token weight, lowers the score by 0.4 points.
So none of the five variants improves on the full recipe: a shorter or longer window lowers the score and so does another loss form or an added regularizer, which leaves the plain loss faded over 10 steps as the setting used in every other experiment.

\textbf{Obs.~\ding{186}: The mask hurts every always-on recipe and barely moves \method{}.}
Masking is a change of a different kind: it removes the massive dimensions of \cref{sec:diagnostics} from the latent loss, so if those dimensions misled the loss, the recipes that keep it on should benefit most.
On validation MATH-500 in \cref{fig:res-ablation}d the opposite happens: the mask lowers OPRD-Bridge from 11.33 to 5.60, OPRD-Bridge + token from 55.57 to 45.25, and the last-layer term kept on beside the token term from 54.75 to 42.00.
\method{} moves 0.4 points, from 59.72 to 59.35.
So masking the latent target hurts all three always-on recipes, while the 10-step crossfade is much less sensitive to the same change.

\begin{figure}[t]
\centering
\includegraphics[width=\textwidth]{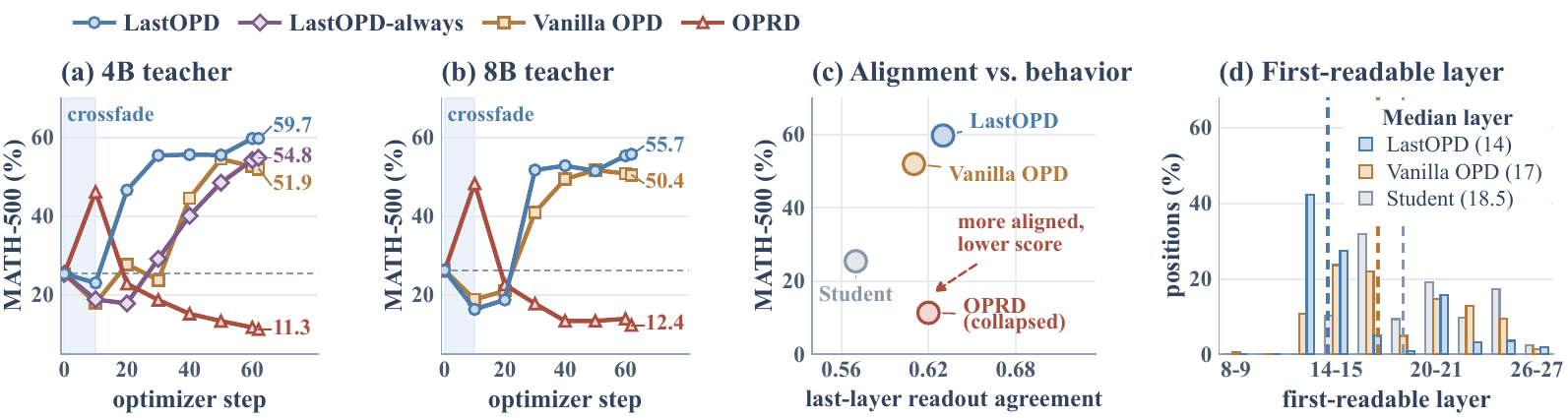}
\vspace{-15pt}
\caption{\textbf{General analysis.} \textbf{(a,~b)}~Validation MATH-500 every 10 steps with the Qwen3-4B and the Qwen3-8B teacher, sharing one legend. \textbf{(c)}~Last-layer readout agreement with the teacher against validation MATH-500. \textbf{(d)}~First layer at which a partial sum enters the top-5 J-Lens readout, over 247 positions from 60 arithmetic chains. Positions never read out are dropped per model.}
\vspace{-5pt}
\label{fig:res-curves}
\end{figure}

\subsection{General Analysis}
\label{sec:res-dynamics}

\cref{fig:res-curves} turns from the final scores to the training run and the trained student.
To see when the gain arrives and whether it stays, \textbf{(a,~b)} track validation MATH-500 every 10 steps under the 4B and the 8B teacher.
To see whether the gain comes with alignment, \textbf{(c)} plots the student's last-layer readout agreement with the teacher against the score.
To see whether the crossfade changed when the student tells, \textbf{(d)} finds the first layer at which a partial sum becomes readable.
We observe:

\textbf{Obs.~\ding{187}: The crossfade accelerates learning and sustains the gain.}
As \cref{fig:res-curves}a shows, \method{} rises from 23.1 to 46.6 on validation MATH-500 in the 10 steps right after the crossfade.
By step 30 it reaches 55.4 and passes the 51.9 that token-only OPD reaches at the end of the 62-step budget, while \method{}-always improves more slowly under continued joint supervision.
In \cref{fig:res-curves}b the jump with the 8B teacher comes 10 steps later, but \method{} again passes the token-only endpoint by step 30 with 51.7 against 50.4.
So with both teachers \method{} reaches the final score of token-only OPD in about half the steps.
The gain also holds past the 62-step budget: in \cref{fig:diag-anatomy}a a \method{} run extended to 150 steps stays near 57 from step 90 onward while OPRD-Bridge remains between 9 and 11.
Together these trajectories show that the early crossfade brings the gain forward and keeps it through the token-only training that follows.
Further trajectories are in Appendix~\ref{app:grpo} and the alignment measurements at the final checkpoints in Appendix~\ref{app:alignment}.

\textbf{Obs.~\ding{188}: Intermediate results become readable earlier under \method{}.}
As \cref{fig:res-curves}c shows, OPRD-Bridge and \method{} end with last-layer readout agreements of 0.62 and 0.63 yet validation MATH-500 scores of 11.33 and 59.72.
To see how intermediate results emerge across depth, we run an arithmetic-chain probe and record the first layer at which the current partial sum enters the top-5 J-Lens readout.
As \cref{fig:res-curves}d shows, among positions where the partial sum becomes readable the median first-readable layer is 14 for \method{} against 18.5 for the untrained student and 17 for token-only OPD.
So the crossfade does not hand the student the teacher's pattern of knowing early and telling late: the final answer surfaces at the same relative depth as before (Appendix~\ref{app:tells}), and the intermediate results now surface earlier.
The student still tells as it goes, which suggests that the latent signal strengthens the student's own way of thinking rather than replacing it with the teacher's.

\section{Conclusion}
\label{sec:conclusion}

This paper studies two failures of latent supervision in on-policy distillation across model sizes: it helps first and harms later, and the alignment it optimizes keeps improving while the student collapses.
Our diagnostics point to a mismatch in how the latent signal is applied: the teacher knows early and tells late while the student tells as it goes, so layers paired by depth do different work and continued alignment may pull the student toward teacher states it cannot understand.
To address this, we introduce \method{}, which applies the latent signal only at the last-layer state and only during a 10-step crossfade into token-level OPD.
Experiments on three teacher--student pairs show that \method{} outperforms existing methods on most datasets and reaches the final score of token-only OPD in about half the steps.
The student still tells as it goes and reaches intermediate results earlier, so the latent signal strengthens the student's way of thinking rather than replacing it with the teacher's.

\bibliography{iclr2027_conference}
\bibliographystyle{iclr2027_conference}

\clearpage

\appendix
\section{Experimental Details}
\label{app:setup}

\paragraph{Models.}
The two Qwen3 teachers and the student differ in shape: Qwen3-4B has 36 layers of width 2560, Qwen3-8B has 36 layers of width 4096, and Qwen3-1.7B-Base has 28 layers of width 2048.
So no student layer has a natural teacher partner, and no latent state can be compared without a projector.
JustRL-1.5B and R1-Distill-1.5B share all 28 layers and widths, which is why we use them as the same-lineage control.
All pairs share a tokenizer and the teacher stays frozen.

\paragraph{Training.}
Each optimizer step draws 32 prompts with 4 responses per prompt at temperature 1.0 and a limit of 8192 tokens, and updates with a learning rate of $10^{-5}$.
On the two cross-size pairs both loss terms carry a base weight of 1, so $\lambda_{\rep}=1$ and the schedule of \cref{eq:crossfade} is the only weighting between them.
On the same-lineage pair the latent term carries a base weight of 1000 as in OPRD~\citep{yang2026oprd}, which brings its step-1 magnitude of 0.026 next to the token term's 0.048, whereas a weight of 1 leaves it near 1/1800 of the token term.
The always-on recipes keep both base weights fixed throughout.
Appendix~\ref{app:latent-weight} shows that raising the weight to 1000 on the 4B pair leaves \method{} nearly unchanged and collapses \method{}-always.
The 62 steps yield about 7.9k rollouts as the training of OPRD~\citep{yang2026oprd}.
For \method{}, teacher latent states are needed only during the first $T_w$ steps, and after that the teacher supplies only its next-token distribution on the student's top-$k$ candidates, which is the cost of standard OPD~\citep{fu2026rethinking}.
Prompts, batches, and optimizer are identical across methods.
Teacher and student both use the Qwen3 chat template with thinking disabled for rollouts and evaluation alike.
To draw trajectories without touching the official protocol, we run a training-time validation on MATH-500 every 10 steps with 8 samples per problem, which tracks the official score within about two points.

\paragraph{Evaluation.}
All scores are measured at the final checkpoint at temperature 0.7 and top-$p$ 0.95, with a 31744-token limit for MATH-500, AIME24, AIME25, AMC23, Minerva, and OlympiadBench, 8192 for AIMO, and 4096 for GSM8K.
MATH-500 and AMC23 use 8 samples per problem, AIME24, AIME25, and AIMO use 16, Minerva and OlympiadBench use 4, and GSM8K uses a single sample.
We never select the best intermediate checkpoint, because a peak can be followed by a collapse and the final checkpoint is what a user would deploy.
MATH-500 also serves the diagnostics of \cref{sec:diagnostics} and the choice of $T_w$, so the other seven datasets are the evidence independent of that choice.

\paragraph{Baselines.}
Token-only OPD is reverse top-16 OPD with no latent term.
OPRD-Bridge (latent-only) aligns all layers by relative depth through a rank-8 bridge frozen during distillation, whose teacher side is the PCA basis of centered teacher states and whose student side is a linear map trained beforehand to match it, and has no token term.
OPRD-Bridge + token adds the token term at constant weight.
On the same-lineage pair the two recipes need no projector and appear as OPRD-Vanilla (latent-only) and OPRD-Vanilla + token.
\method{}-always (no fade) keeps both terms of \method{} at their base weights throughout the training process and differs from \method{} only in the schedule.

\section{Protocols Behind the Diagnostic Figure}
\label{app:diag}

This appendix gives the protocol behind each panel of \cref{fig:diag-anatomy}.

\subsection{Long runs (panel a)}
\label{app:proto-long}
Both runs use the training recipe of \cref{sec:setup} and continue to 150 optimizer steps instead of 62, with every other setting unchanged.
The curves are the training-time validation scores on MATH-500, computed every 10 steps with 8 samples per problem.

\subsection{Layerwise readouts (panel b)}
\label{app:proto-readout}
We use J-Lens~\citep{gurnee2026verbalizable}, a lens fitted per model that decodes the latent state of every layer into next-token logits.
The prompts are ten one-digit expressions such as ``$3*2+2=$'' after a two-shot prefix, chosen so that the intermediate product and the answer are single tokens that differ from every operand.
At the final position we decode the top-1 token at every layer: green marks the final answer, orange the intermediate product, and grey anything else.
Panel b shows one expression for the untrained student, the teacher, OPRD-Bridge~(62 steps), and token-only OPD~(62 steps).
All of them are shown in Appendix~\ref{app:tells}.

\subsection{Readout agreement (panel c, solid lines)}
\label{app:proto-agreement}
The solid curves ask at which layer a model has settled on the token it will emit.
We take 64 MATH-500 problems and three mid-generation positions in each, 192 positions in all, and decode the top-1 J-Lens readout at every layer.
A layer counts as settled when that readout already equals the token the model finally emits, and the curve is the fraction of settled positions against relative depth.

\subsection{Value probe (panel c, dashed lines)}
\label{app:proto-probe}
The dashed curves ask at which layer a model holds a value it has not yet said.
We build all 729 expressions ``$a+b+c=$'' with $a,b,c$ from 1 to 9, prepend the two-shot prefix ``2+2=4'' and ``7+1=8'' to ensure one-token outputs, and take every layer's latent state at the ``$=$'' position.
The target is the partial sum $a+b$, which takes 17 values from 2 to 18, so chance is 6\%.
For each layer we standardize the states with the training-set mean and deviation and fit a bias-free linear classifier by cross-entropy, using Adam at learning rate 0.01 with weight decay $10^{-3}$ for 300 steps.
The 729 expressions are split into 583 for training and 146 for testing, and the curve is test accuracy against depth.

\subsection{Layer-to-layer CKA (panels d and e)}
\label{app:proto-cka}
We compute linear CKA between the latent states of every student layer and every teacher layer on 200 on-policy texts of up to 512 tokens drawn from the same pool as the training rollouts.
The token positions are sampled once and shared across all layers and models, so every cell of the map sees the same tokens.
Panel e repeats the computation after removing each model's top-1\% massive channels.

\subsection{Massive activations}
\label{app:proto-massive}
For each of the five models in this paper we ran 64 on-policy texts of 512 tokens through the network and averaged the absolute activation of every hidden channel over all layers and tokens.
The top-1\% channels are the massive dimensions, and \cref{tab:massive} lists their count, the median channel, the largest channel, and the ratio between the two.
Every model has a largest channel at 32 to 83 times its median, and the two same-lineage models share 93\% of their top channels.

\begin{table}[t]
\small
\caption{Massive activations in the five models used in this paper: mean absolute activation per hidden channel over 64 on-policy texts of 512 tokens, all layers and tokens. The top-1\% channels are the massive dimensions removed in panel e of \cref{fig:diag-anatomy}.}
\label{tab:massive}
\begin{center}
\begin{tabular}{lccccc}
\toprule
Model & hidden $d$ & top-1\% channels & median channel & largest channel & largest/median \\
\midrule
Qwen3-1.7B-Base & 2048 & 20 & 159.6 & 7567.0 & 47.4 \\
Qwen3-4B & 2560 & 25 & 36.4 & 2079.3 & 57.2 \\
Qwen3-8B & 4096 & 40 & 53.6 & 4448.4 & 83.0 \\
JustRL-1.5B & 1536 & 15 & 48.0 & 1532.2 & 31.9 \\
R1-Distill-1.5B & 1536 & 15 & 45.6 & 1818.3 & 39.9 \\
\bottomrule
\end{tabular}
\end{center}
\end{table}

\section{Additional Results}
\label{app:more}

\subsection{Full ablation table}
\label{app:ablation-table}
\cref{tab:ablation} lists every run behind \cref{fig:res-ablation}.
\begin{table}[t]
\caption{Ablations with the Qwen3-4B teacher, official MATH-500 at the final checkpoint. Each block changes one thing about \method{}. The last column is the gap to the full method~(58.95). The reversed-projector, Gram-matrix, and repetition-penalty rows also change the schedule, keeping both terms on throughout, so their like-for-like reference is the always-on row~(53.77). The repetition-penalty row comes from the earlier regime~(penalty 1.05). In the masked block both columns are validation MATH-500 and the last column is masked minus unmasked.}
\label{tab:ablation}

\begin{center}
\scriptsize
\setlength{\tabcolsep}{4pt}
\renewcommand{\arraystretch}{1.12}
\resizebox{\textwidth}{!}{%
\begin{tabular}{lcclcc}
\toprule
Configuration & Latent term & Token term & Schedule & MATH-500 & Gap \\
\midrule
\rowcolor{bandA}\multicolumn{6}{c}{\textbf{Schedule: when the latent term is on}} \\
\method{}~(full) & last layer & reverse top-16 & latent $1\!\to\!0$, token $0\!\to\!1$ over 10 steps & \cellcolor{bestcell}\textbf{58.95} & --- \\
hard switch & last layer & reverse top-16 & latent only for 10 steps, then token only & 50.45 & \textcolor{dneg}{$-$8.50} \\
fade only & last layer & reverse top-16 & latent $1\!\to\!0$, token constant & 50.90 & \textcolor{dneg}{$-$8.05} \\
raise token only & --- & reverse top-16 & no latent term, token $0\!\to\!1$ & 54.15 & \textcolor{dneg}{$-$4.80} \\
raise token, hold latent 10 steps & last layer & reverse top-16 & latent constant for 10 steps then off, token $0\!\to\!1$ & 55.73 & \textcolor{dneg}{$-$3.22} \\
always on & last layer & reverse top-16 & both constant & 53.77 & \textcolor{dneg}{$-$5.18} \\
token-only OPD & --- & reverse top-16 & token constant & 53.40 & \textcolor{dneg}{$-$5.55} \\
\midrule
\rowcolor{bandB}\multicolumn{6}{c}{\textbf{Crossfade window}} \\
5 steps & last layer & reverse top-16 & crossfade over 5 steps & 54.62 & \textcolor{dneg}{$-$4.33} \\
10 steps~(full) & last layer & reverse top-16 & crossfade over 10 steps & \cellcolor{bestcell}\textbf{58.95} & --- \\
15 steps & last layer & reverse top-16 & crossfade over 15 steps & 54.47 & \textcolor{dneg}{$-$4.48} \\
\midrule
\rowcolor{bandC}\multicolumn{6}{c}{\textbf{Extra terms and loss form}} \\
+ KL to the initial policy & last layer & reverse top-16 & crossfade, KL rises with token & 58.55 & \textcolor{dneg}{$-$0.40} \\
+ massive-activation mask & last layer, top-1\% dims removed & reverse top-16 & crossfade & 58.07 & \textcolor{dneg}{$-$0.88} \\
projector direction reversed & last layer, teacher$\to$student & reverse top-16 & both constant & 50.00 & \textcolor{dneg}{$-$8.95} \\
Gram-matrix structural loss & last layer, Gram & reverse top-16 & both constant & 46.23 & \textcolor{dneg}{$-$12.72} \\
repetition penalty 1.05~(earlier regime) & last layer & reverse top-16 & both constant & 44.62 & \textcolor{dneg}{$-$14.33} \\
\midrule
\rowcolor{bandD}\multicolumn{6}{c}{\textbf{Massive-activation mask applied to other recipes and \method{}}} \\
OPRD-Bridge & all layers & --- & always on & 11.33 $\rightarrow$ 5.60 & \textcolor{dneg}{$-$5.73} \\
OPRD-Bridge + token & all layers & reverse top-16 & always on & 55.57 $\rightarrow$ 45.25 & \textcolor{dneg}{$-$10.32} \\
\method{}-always & last layer & reverse top-16 & always on & 54.75 $\rightarrow$ 42.00 & \textcolor{dneg}{$-$12.75} \\
latent-only, trainable head & all layers & --- & always on & 0.40 $\rightarrow$ 0.05 & \textcolor{dneg}{$-$0.35} \\
\method{}~(full) & last layer & reverse top-16 & crossfade over 10 steps & 59.72 $\rightarrow$ 59.35 & \textcolor{dneg}{$-$0.37} \\
\bottomrule
\end{tabular}}
\end{center}
\end{table}

\begin{figure}[t]
\centering
\includegraphics[width=0.5\textwidth]{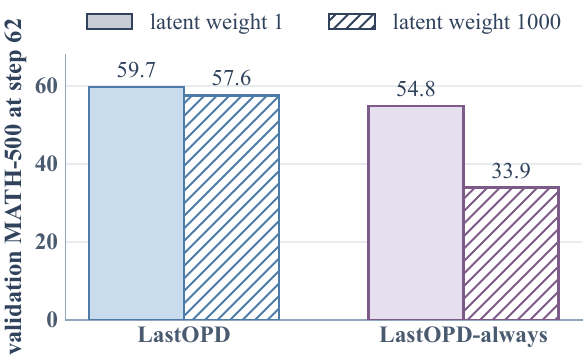}
\caption{\textbf{Latent weight 1 against 1000 on the Qwen3-4B pair.} Validation MATH-500 at step 62 for \method{} and \method{}-always. Solid: weight 1. Hatched: weight 1000.}
\label{fig:app-weight}
\end{figure}
\subsection{Raising the latent weight on the 4B pair}
\label{app:latent-weight}
The two cross-architecture pairs train with a latent weight of 1 and the same-architecture pair with 1000, following OPRD~\citep{yang2026oprd}.
To check that the different latent weights of Appendix~\ref{app:setup} are not what separates the recipes, we reran \method{} and \method{}-always on the Qwen3-4B pair with the latent weight raised from 1 to 1000.
As \cref{fig:app-weight} shows, \method{} ends at 57.57 against 59.72 at weight 1.
So the weight of 1000 brings no gain to \method{}.
\method{}-always instead falls from 25.30 at step 10 to 13.33 by step 20 on validation MATH-500 and recovers only to 33.90 against 54.75.
So a stronger latent term that stays on brings back the collapse of \cref{sec:diagnostics}, and only the fade keeps the gain.
Therefore the results above suggest that a weight of 1 suffices on a cross-architecture pair and that only the crossfade tolerates a larger one.

\begin{figure}[t]
\centering
\includegraphics[width=0.55\textwidth]{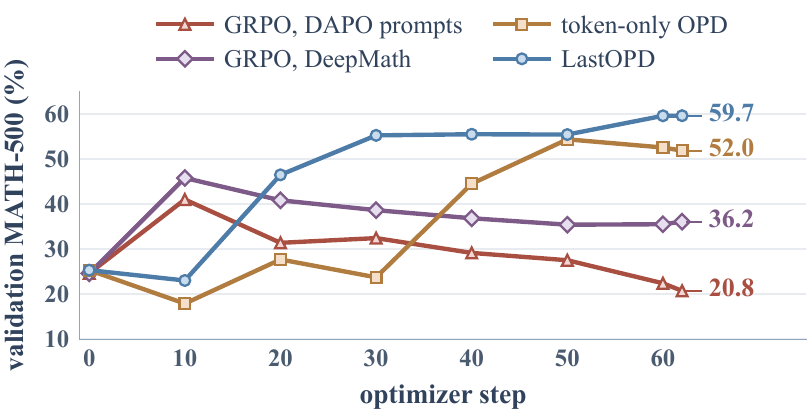}
\caption{\textbf{Reinforcement learning without a teacher.} Training-time validation MATH-500 every 10 steps for GRPO on the DAPO prompts and on DeepMath, against token-only OPD and \method{} with the Qwen3-4B teacher.}
\label{fig:app-grpo}
\end{figure}

\subsection{Reinforcement learning without a teacher}
\label{app:grpo}
To check that the collapse is not what happens to this student under any on-policy training, we ran GRPO on the same prompts with no teacher.
As \cref{fig:app-grpo} shows, it peaks at 41.1 at step 10 and decays to 20.8, and a GRPO run on DeepMath peaks at 45.9 and decays to 36.2.
So a student trained without a teacher also drifts, whereas every distillation run with a token term ends above where it started.

\subsection{Alignment at the final checkpoints}
\label{app:alignment}
To check that the gain of \method{} is not a gain in alignment, we measure the last-layer CKA between student and teacher at the final checkpoints without any projector and with the massive dimensions removed.
It is 0.570 for token-only OPD and 0.546 for \method{}, so the student that scores 5.55 points higher is slightly less aligned.
The collapsed OPRD-Bridge student ends with a last-layer readout agreement of 0.62, within 0.01 of \method{} at 0.63, yet scores 12.12 against 58.95.
Alignment therefore tracks the objective that was optimized and not the behavior that resulted, which is the second failure of \cref{sec:diagnostics} seen again at the end of training.

\subsection{Two collapsed answers side by side}
\label{app:collapse-looks}
To see what the collapsed student actually writes, we read its answers next to those of the \method{} student on the same prompts.
The collapsed student keeps the teacher's form and loses the arithmetic.
Asked for the greatest common divisor of 128 and 144, it writes a well-formed Euclidean algorithm with step headings and boxed formatting, and then divides 128 by 16 to obtain a quotient of 1 with remainder 0.
Asked to convert $(0,3)$ to polar coordinates, it writes $\sqrt{0^2+3^2}=\sqrt{0+3}=\sqrt3$ and then reports $(3,\pi/2)$ anyway.
On other problems it writes $4+3=10$ inside an otherwise fluent derivation.
The \method{} student answers both prompts correctly with the teacher's structure, stating the algorithm before applying it, and it still reads out single-digit products and sums correctly in nine of ten one-step probes.
So the damage is selective: one-step arithmetic and the surface form of a proof survive, while the composition of steps, carries, and remainders is lost.

\begin{figure}[t]
\centering
\includegraphics[width=0.72\textwidth]{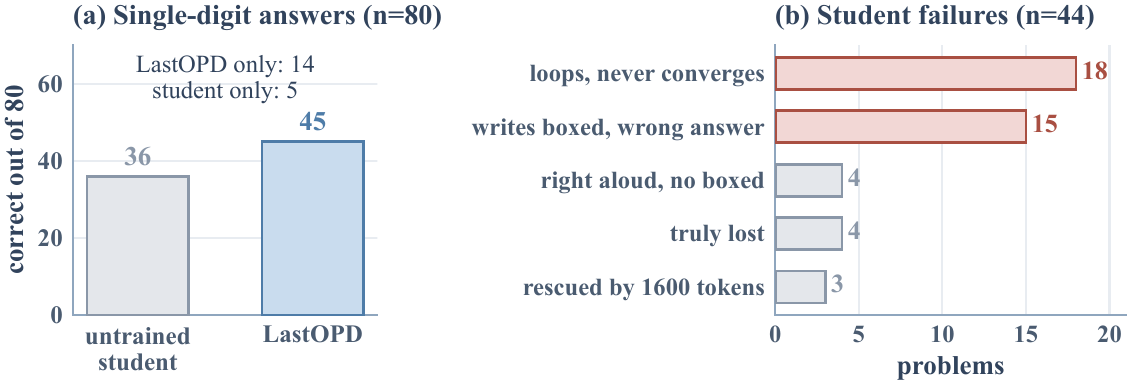}
\caption{\textbf{Case-study statistics.} (a)~Greedy accuracy of the untrained student and \method{} on the 80 MATH-500 problems with a single-digit answer, under the same chat prompt with a boxed instruction. (b)~How the untrained student's 44 failures break down after a 1600-token retry.}
\label{fig:app-cases}
\end{figure}

\begin{figure}[t]
\centering
\includegraphics[width=0.9\textwidth]{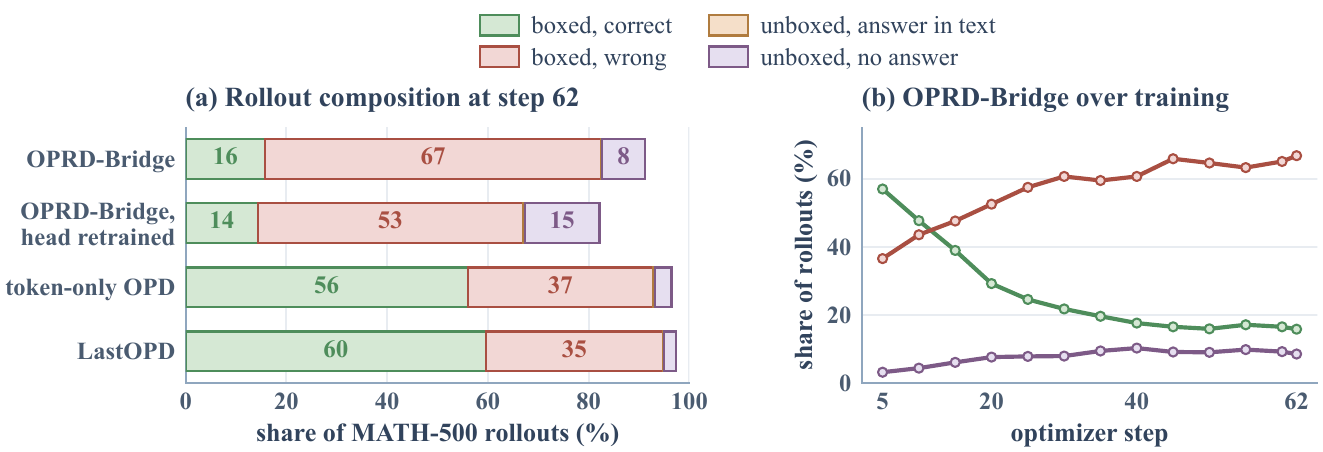}
\caption{\textbf{Four kinds of MATH-500 rollout.} \textbf{(a)}~Share of rollouts of each kind at step 62 for a replicate OPRD-Bridge run, the same run with only its LM head retrained on teacher text, token-only OPD, and \method{}. Problems whose answer is one or two characters are left out of the unboxed kinds. \textbf{(b)}~The same shares along the OPRD-Bridge run.}
\label{fig:app-fourclass}
\end{figure}

\subsection{What the collapsed student says}
\label{app:collapsed-says}
A few cases do not make a pattern, so we turn to all MATH-500 rollouts of the collapsed student.
\cref{fig:app-fourclass} sorts them into four kinds: boxed and correct, boxed and wrong, unboxed with the correct answer somewhere in the text, and unboxed without it.
At step 62 the shares are 15.8\%, 66.8\%, 0.2\%, and 8.5\%, so two thirds of the rollouts end in a wrong boxed answer and almost none stop short of boxing one.
These wrong answers are also short, 765 tokens on average against a cap of 8192, whereas the 8.5\% without an answer are arithmetic loops that run to the cap.
Along the run the wrong share rises from 36.5\% at step 5 to 66.8\% at step 62 while the correct share falls from 57.0\% to 15.8\%, so the collapse is a steady conversion of right answers into boxed wrong ones.
Retraining only the LM head on teacher text does not reverse it: rollouts move from wrong to unboxed and not to correct.
For \method{} and token-only OPD at step 62 the correct shares are 59.6\% and 56.0\% and the loops 2.4\% and 3.3\%.

\subsection{Where the gain shows up in the outputs}
\label{app:gain-outputs}
To see where the gain of \method{} over the untrained student comes from, we take the 80 MATH-500 problems whose answer is a single digit, because a one-digit answer can be checked without a parser and cannot be guessed from the format.
As \cref{fig:app-cases} shows, \method{} answers 45 correctly against 36 for the untrained student, winning 14 problems and losing 5, and all 45 of its correct answers stop within 600 tokens.
The untrained student mostly fails by looping without converging or by boxing a wrong answer.
To rule out truncation as the cause, we retry its 44 failures with a 1600-token budget, and only 3 are rescued.

\begin{figure}[t]
\centering
\includegraphics[width=\textwidth]{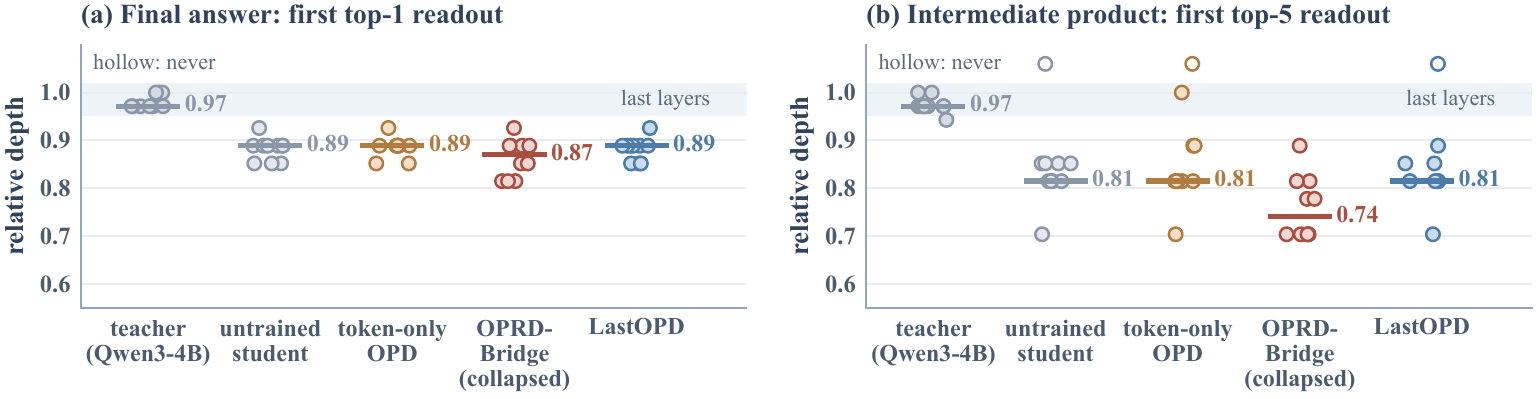}
\caption{\textbf{When the answer becomes readable, on ten one-digit prompts.} Each dot is one prompt and the bar is the median. (a)~Depth at which the final answer first becomes the top-1 J-Lens readout. (b)~Depth at which the intermediate product first enters the top-5 readout. Hollow dots: never.}
\label{fig:app-readout}
\end{figure}

\begin{figure}[t]
\centering
\includegraphics[width=0.58\textwidth]{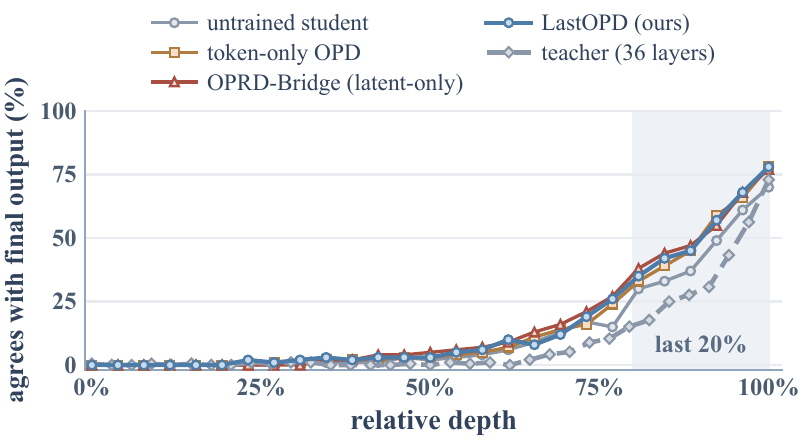}
\caption{\textbf{Readout agreement with the final output by relative depth}, for the untrained student, the three trained students, and the teacher.}
\label{fig:app-cliff}
\end{figure}

\subsection{Training does not change when the student tells}
\label{app:tells}
\begin{figure}[p]
\centering
\includegraphics[width=\textwidth]{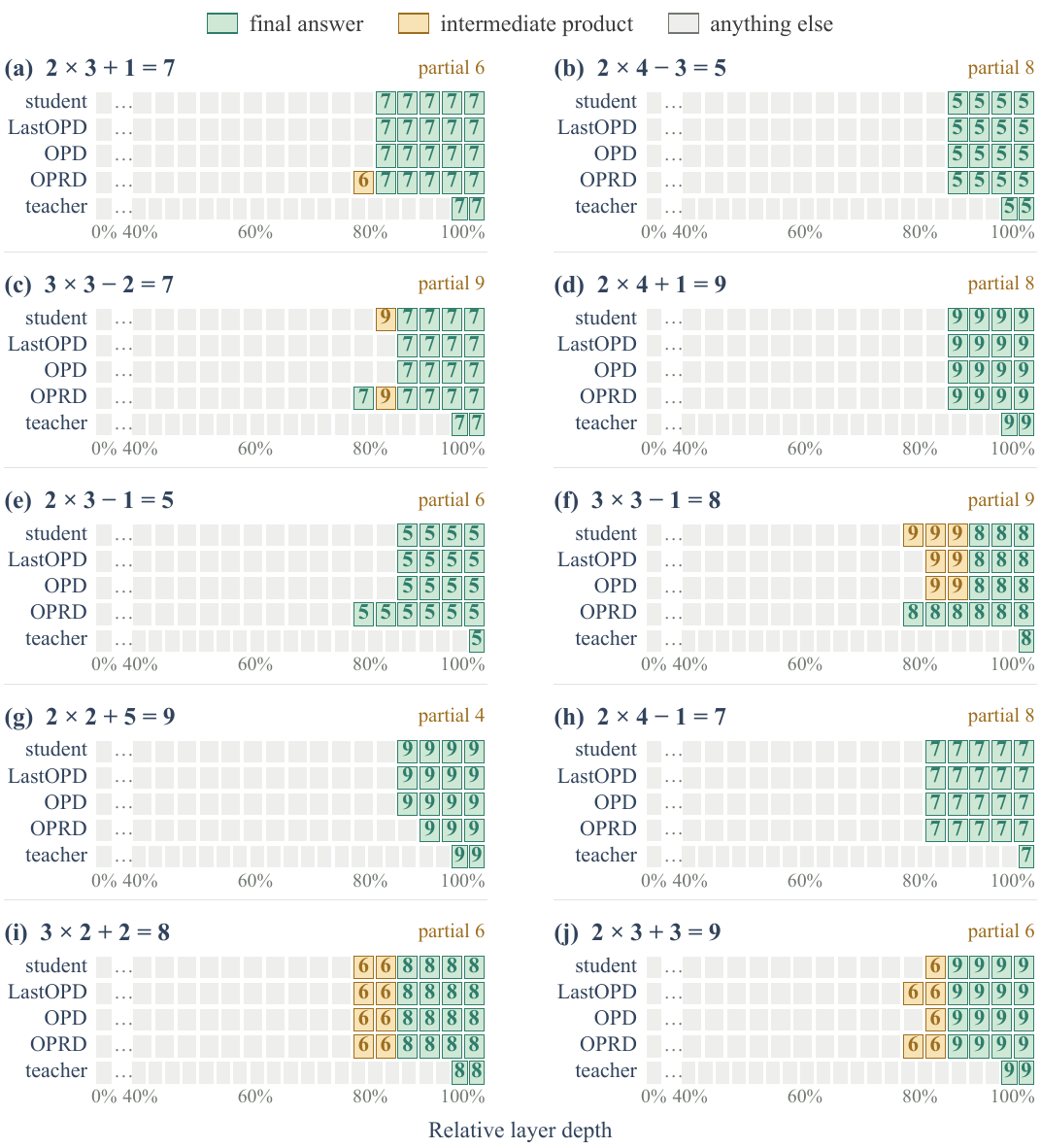}
\caption{\textbf{Layerwise J-Lens readout on all ten one-digit prompts.} Each panel repeats \cref{fig:diag-anatomy}b for one prompt and five models: the untrained student, \method{}, token-only OPD, OPRD-Bridge, and the teacher. Depth is normalized per model, the 0--40\% interval is compressed to the first layer, and green marks the final answer, amber the intermediate product, and grey anything else.}
\label{fig:app-readout-ten}
\end{figure}

To check whether any recipe changes when the student says its answer, \cref{fig:app-readout} repeats the readout of \cref{fig:diag-anatomy}b on ten one-digit prompts.
The teacher's answer becomes readable only in its last layers, at a median relative depth of 0.97, while the untrained student, token-only OPD, OPRD-Bridge, and \method{} all tell at a median of 0.87 to 0.89.
The intermediate product shows the same split, with the teacher at 0.97 and the students between 0.74 and 0.81.
Whatever the recipe, training leaves the depth at which the student tells unchanged.
This holds for the final answer, whereas the partial sums of \cref{fig:res-curves}d are another readout and move earlier.
\cref{fig:app-cliff} shows the same picture over 192 positions: the four students share one curve, near zero until 60\% depth and rising to 70--78\% at the last layer, and every recipe lifts it by the same few points while the teacher stays later.
\cref{fig:app-readout-ten} shows the full readout of every prompt: all four students surface the answer from about 85\% depth on, whereas the teacher surfaces it only in its last one or two layers, and the intermediate product appears in the students' top-1 readouts on five of the ten prompts but never in the teacher's.

\begin{figure}[t]
\centering
\includegraphics[width=\textwidth]{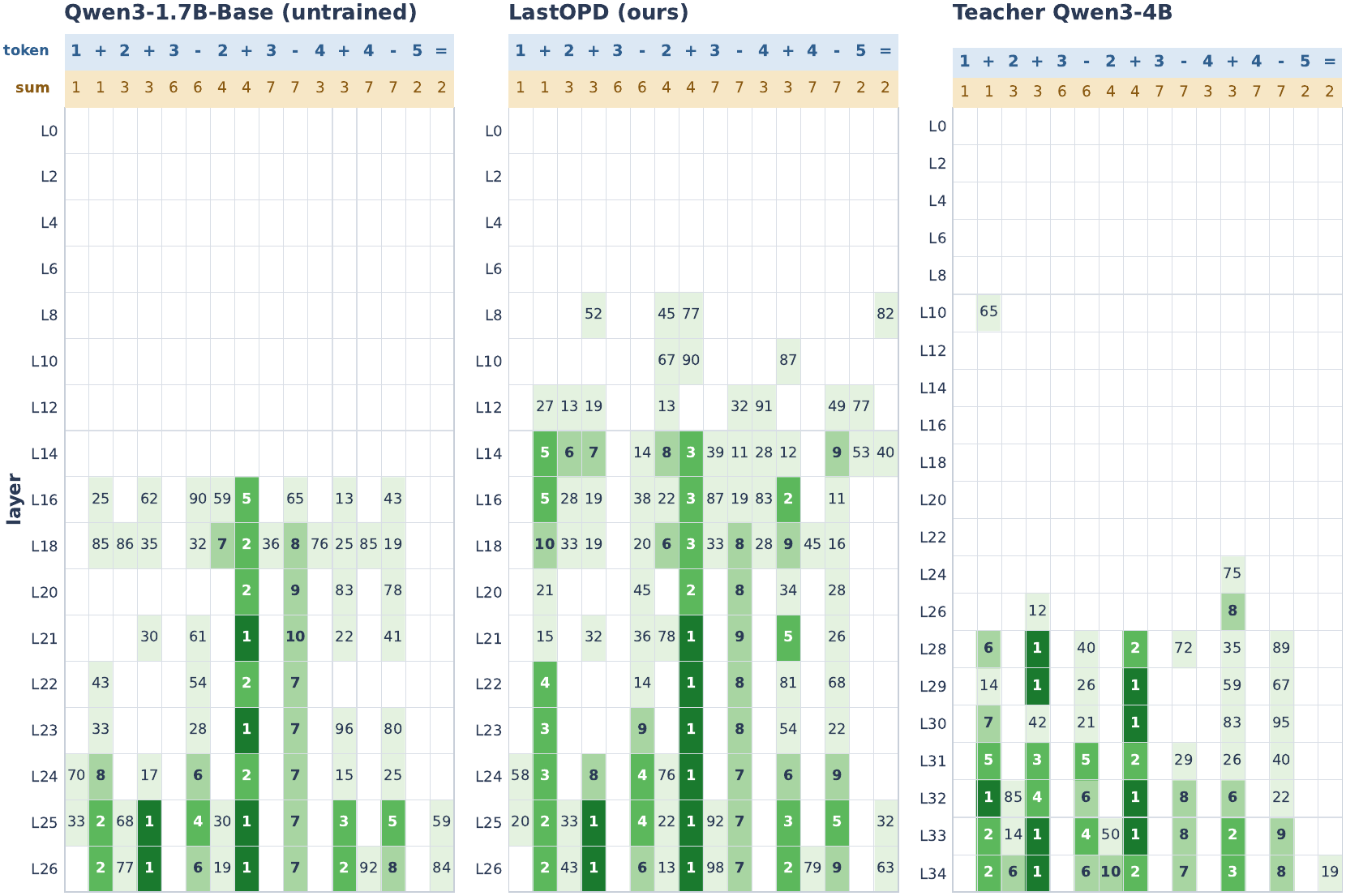}
\caption{\textbf{Reading the running partial sum along one chain.} For $1+2+3-2+3-4+4-5=$ the top row of each panel lists the token read in and, below it, the partial sum at that token. Each cell is the rank of that partial sum in the layer's J-Lens readout, blank when it falls outside the top 99. The untrained student and \method{} read the running value out from the middle layers on, while the teacher shows little before its last ten layers.}
\label{fig:app-chain-readout}
\end{figure}

\subsection{Where the running value surfaces along one chain}
\label{app:chain}
\cref{fig:app-chain-readout} follows a single chain token by token and asks, at every layer, how highly the current partial sum ranks in the readout.
Both students carry the running value in the readable space from layers 14 to 16 on, and \method{} brings it to rank 1 at more positions and from slightly shallower layers than the untrained student.
The teacher shows only a faint trace of the value before layer 26 of 36 and reads it out at rank 1 from layer 28 across most of the chain.

\begin{figure}[t]
\centering
\includegraphics[width=0.72\textwidth]{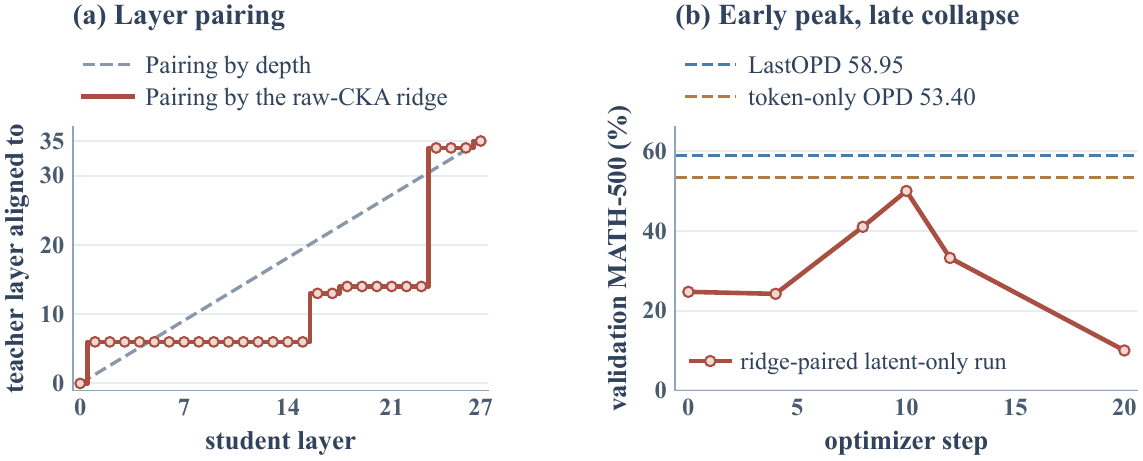}
\caption{\textbf{Pairing layers by the raw CKA ridge.} (a)~The layer map used by the run, against pairing by depth. (b)~Its validation trajectory against the token-only OPD and \method{} final scores.}
\label{fig:app-ridge}
\end{figure}

\subsection{Pairing layers by the raw CKA ridge}
\label{app:ridge}
Before the massive activations were identified, we tested whether a better layer map alone could rescue layerwise latent distillation.
Each student layer was paired with the teacher layer that the raw CKA map marks as most similar.
As \cref{fig:app-ridge}a shows, this sends student layers 1 to 15 to teacher layer 6 and the last four student layers to teacher layers 34 and 35, and the run kept every other setting of OPRD-Bridge.
As \cref{fig:app-ridge}b shows, it followed the same lifecycle as pairing by depth and peaked near 50 at step 10 before collapsing by step 20.
This is evidence that this map did not help.

\subsection{Fading the latent term over the whole run}
\label{app:slowfade}
To check that a short window is what matters and not the fade itself, we ran a variant that fades the latent weight to zero over all 62 steps instead of 10.
It ended at 51.65 on validation MATH-500, at the level of token-only OPD, so a fade that keeps the latent term on for most of the run gives no gain.

\end{document}